%% file: main.tex
\documentclass[10pt,twocolumn,letterpaper]{article}

\usepackage[accsupp]{axessibility}

\usepackage{iccv}
\usepackage{times}
\usepackage{epsfig}
\usepackage{graphicx}
\usepackage{amsmath}
\usepackage{amssymb}
\usepackage[dvipsnames,table]{xcolor}
\usepackage{gensymb}
\usepackage{multirow}
\usepackage{rotating}
\usepackage{float}
\usepackage{threeparttable}
\usepackage{tabularx}
\usepackage{array}
\usepackage[normalem]{ulem}
\usepackage{xpatch}
\usepackage{booktabs}
\usepackage{siunitx}
\usepackage{enumitem}

\usepackage{soul}
\usepackage{changepage,threeparttable}

\usepackage{tikz}
\def\checkmark{\tikz\fill[scale=0.35](0,.35) -- (.25,0) -- (1,.7) -- (.25,.15) -- cycle;}

\usepackage[breaklinks=true,bookmarks=false]{hyperref}

\iccvfinalcopy
\def\iccvPaperID{11094}

\ificcvfinal\fi

\makeatletter
\xpatchcmd{\paragraph}{3.25ex \@plus1ex \@minus.2ex}{1pt plus 1pt minus 1pt}{\typeout{success!}}{\typeout{failure!}}
\makeatother

\begin{document}

\title{MemorySeg: Online LiDAR Semantic Segmentation with a Latent Memory}

\author{
Enxu Li
\quad Sergio Casas
\quad Raquel Urtasun \\
Waabi
\quad University of Toronto \\
{\texttt{\{tli, sergio, urtasun\}@waabi.ai}}
}

\maketitle
\ificcvfinal\thispagestyle{empty}\fi

\newcommand*{\proposed}{\textsc{MemorySeg}}

\begin{abstract}
Semantic segmentation of LiDAR point clouds has been widely studied in recent years, with most existing methods focusing on tackling this task using a single scan of the environment. 
However, leveraging the temporal stream of observations can provide very rich contextual information on regions of the scene with poor visibility (e.g., occlusions) or sparse observations (e.g., at long range), and can help reduce redundant computation frame after frame. In this paper, we tackle the challenge of exploiting the information from the past frames to improve the predictions of the current frame in an online fashion. To address this challenge, we propose a novel framework for semantic segmentation of a temporal sequence of LiDAR point clouds that utilizes a memory network to store, update and retrieve past information. Our framework also includes a novel regularizer that penalizes prediction variations in the neighborhood of the point cloud. Prior works have attempted to incorporate memory in range view representations for semantic segmentation, but these methods fail to handle occlusions and the range view representation of the scene changes drastically as agents nearby move. Our proposed framework overcomes these limitations by building a sparse 3D latent representation of the surroundings.
We evaluate our method on SemanticKITTI, nuScenes, and PandaSet. Our experiments demonstrate the effectiveness of the proposed framework compared to the state-of-the-art. For more information, visit the project website: \url{https://waabi.ai/research/memoryseg}.
\end{abstract}

\input{sections/intro}
\input{sections/related_work}
\input{sections/method}
\input{sections/experiments}
\input{sections/conclusion}

{\small
\bibliographystyle{ieee_fullname}
\bibliography{memseg}
}

\clearpage
\section*{Supplementary Materials}
  \addcontentsline{toc}{section}{Appendix}
  \setcounter{section}{0}
  \renewcommand{\thesection}{\Alph{section}}

  In this supplementary material, we first describe the implementation details including the Memory Refinement Module (in Sec.~\ref{sec:mrm}) and the added motion header (in Sec.~\ref{sec:motion}) for separating moving and static actors in SemanticKITTI \cite{behley2019semantickitti}. Subsequently, we show the addition results of the following:
\begin{itemize}
   \item \proposed \ compared against state-of-the-art methods on test and validation set of SemanticKITTI \cite{behley2019semantickitti} single-scan benchmark in Sec.~\ref{sec:sk_single}
   \item \proposed \ compared against our baseline on validation set of SemanticKITTI \cite{behley2019semantickitti} multi-scan benchmark in Sec.~\ref{sec:sk_multi}
   \item \proposed \ compared against state-of-the-art and our baselines on validation set of nuScenes \cite{caesar2020nuscenes} in Sec.~\ref{sec:nusc_val}
   \item ablation analysis of memory voxel size, padding neighbourhood size, instance cutMix and test-time augmentation in Sec.~\ref{sec:ablation}
   \item visualization of the latent memory in Sec.~\ref{sec:mem_vis}
   \item additional qualitative results of \proposed \ compared with our baseline in Sec.~\ref{sec:qua}
\end{itemize}
\input{sections/supp_details}
\input{sections/supp_results}

\end{document}

%% file: sections/intro.tex
\section{Introduction}
\begin{figure}[t]
    \centering
    \includegraphics[width=\linewidth]{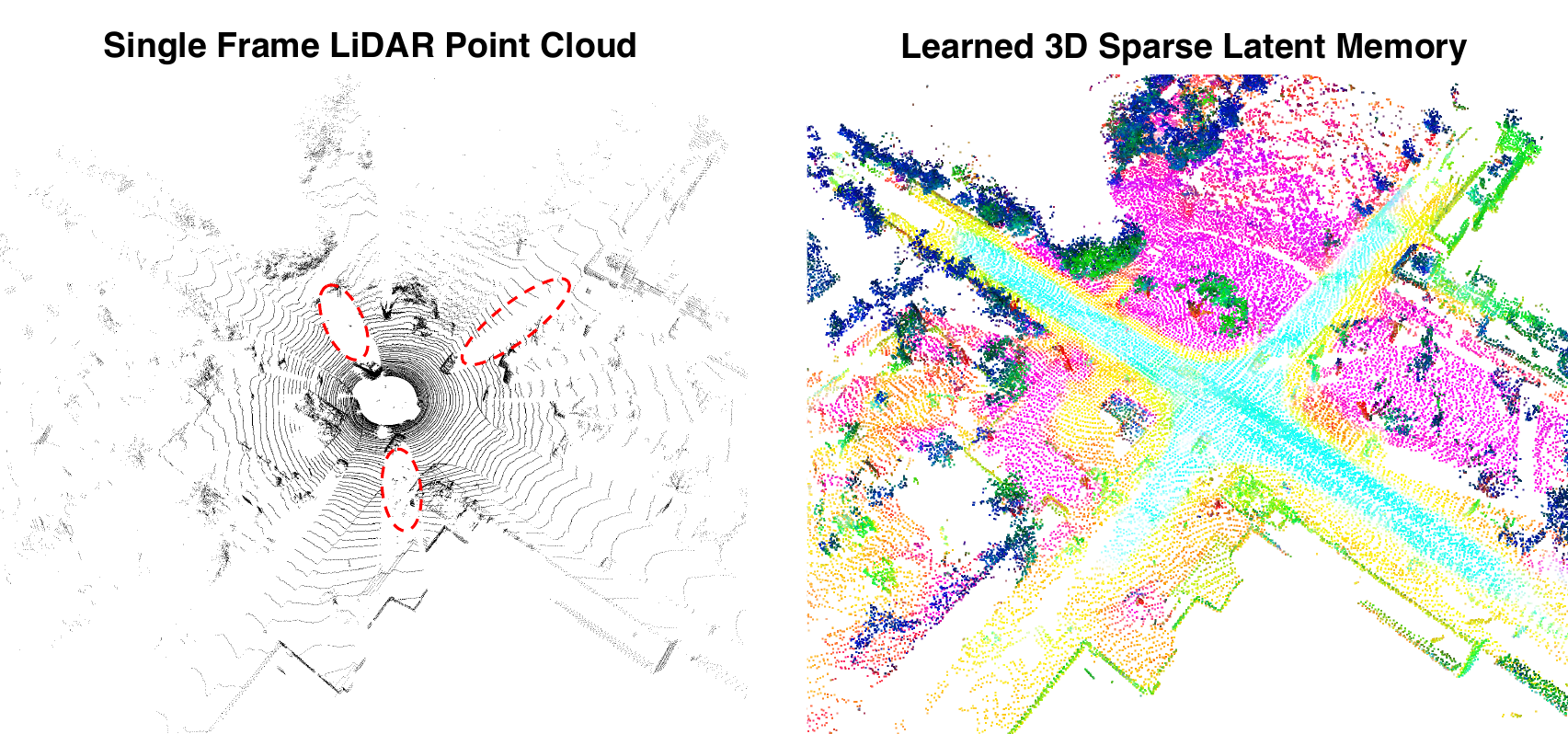}
    \caption{
        Objects could be partially occluded in single frame LiDAR point cloud. Our approach learns a 3D latent memory representation for better contextualizing the online observations. We apply PCA \cite{jollieffe2005pca} to reduce the latent dimension to 3 and plot as RGB. (Best viewed in color and zoomed-in.)
    }
    \label{fig:memory_comparison}
    \vspace{-10px}
\end{figure}

Semantic segmentation of LiDAR point clouds is a key component for the safe deployment of self-driving vehicles (SDV).
It enables SDVs to enhance their understanding of the surrounding environment by categorizing every 3D point into specific classes of interest, such as vehicles, pedestrians, traffic signs, buildings, roadways, etc.
This rich and precise 3D representation of the environment can then be used for various applications such as generating online or offline semantic maps, building localization priors, or making the shapes of an object tracker more precise. 

LiDAR data is typically captured as a continuous stream of data, where every fraction of a second (typically 100ms) a new point cloud is available. 
Despite this fact, most LiDAR  segmentation approaches process each frame independently \cite{milioto2019rangenet++,zhang2020polarnet,xu2020squeezesegv3,tang2020searching,zhu2021cylindrical,cheng20212,xu2021rpvnet} due to the computational and memory complexity associated with 
processing large amounts of 3D point cloud data. 
However, reasoning about a single frame suffers from the sparsity of the observations, particularly at range, and has difficulty handling occluded objects. Furthermore, the absence of motion information can make categorizing certain objects difficult.

Several approaches \cite{shi2020spsequencenet,schuett2022abstractflow} utilize a sliding window approach, where a small set of past frames is processed independently at every time step. The main shortcoming of this approach is its limited temporal context (typically under 1 second) due to resource constraints, as processing multiple LiDAR scans at once is expensive.
TemporalLidarSeg \cite{duerr2020lidar} proposed to utilize a latent spatial memory to retain information about the past while avoiding redundant computation every time a new scan is available, in a range view (RV) representation. 
However, the RV of the scene changes drastically as the SDV or the other actors move due to changes in perspective. 
As a result, the memory can be dominated by nearby objects, which appear larger in the RV representation, making it challenging to be updated from frame to frame.
Occlusion can be particularly challenging as the occluder now occupies the same spatial region that was previously describing the occluded object. This limits the usefulness of the memory.
In contrast to RV, 3D is a metric space where the distances between points are preserved regardless of the viewpoint or relative distance to the SDV. Thus, representing the memory in 3D enables learning size priors for different classes.
It allocates equal representation power to them regardless of the distance to the SDV, and  enables easy understanding of motion. Moreover, previously observed regions that are currently occluded can be remembered in the memory as the occluder and occluded objects occupy different 3D regions, even though they share the same space in RV. 
Despite these advantages, 3D memory has been overlooked in LiDAR semantic segmentation.

In this paper we propose \proposed, a novel online LiDAR segmentation model  that 
recurrently updates a sparse 3D latent memory as new observations are received, efficiently and effectively accumulating evidence from past observations. 
Fig.~\ref{fig:memory_comparison} illustrates the expressive power of such memory.
In a single scan, objects are hard to identify due to sparsity (particularly at range) and lack of semantics, and occluded areas have no observations. In contrast, our latent memory is much denser, providing a rich context to separate different classes, especially in currently occluded regions.

To achieve an effective memory update that takes into account both the past and the present for accurate decoding, our method addresses several challenges.
First, we align the previous memory with the current observation in the latest ego frame, to compensate for the SDV motion. 
Second, the sparsity level of the memory and the observation embeddings are different, making fusion non-trivial.
The latent memory is denser, and thus some currently unobserved locations may be present in the memory.
Complementarily, some regions in the current scan have not been observed before, such as new observations appearing as the SDV drives further.
For this reason, we propose a mechanism to fill in missing regions in the current observations and add new observations to the memory during the update. 
Third, other objects are also moving (potentially in opposite direction), thus fusing the memory and observation embeddings requires a large receptive field. We address this via a carefully designed architecture that achieves a large receptive field yet preserves sparsity for efficiency.
Aditionally, we introduce a novel point-level neighbourhood variation regularizer which penalizes significant differences in semantic predictions within local 3D neighborhoods. 

Extensive experiments in SemanticKITTI \cite{behley2019semantickitti}, nuScenes \cite{caesar2020nuscenes} and PandaSet \cite{xiao2021pandaset} demonstrate that \proposed \ outperforms current state-of-the-art semantic segmentation methods that rely purely on LiDAR on multiple benchmarks.

%% file: sections/related_work.tex
\section{Related Work}

In this section, we start by reviewing LiDAR semantic segmentation approaches in the literature and then focus on methods proposed for temporal LiDAR reasoning.
\subsection{LiDAR Semantic Segmentation}

LiDAR-based semantic segmentation methods can be classified into four categories based on the type of data being processed: point-based \cite{qi2017pointnet,qi2017pointnet++,hu2020randla,thomas2019kpconv}, projection-based \cite{xu2020squeezesegv3,cortinhal2020salsanext,zhang2020polarnet}, voxel-based \cite{cheng20212,zhu2021cylindrical,hou2022pvkd}, or a combintation of the data representations \cite{tang2020searching,xu2021rpvnet}. 

Point-based approaches \cite{qi2017pointnet,qi2017pointnet++,hu2020randla,thomas2019kpconv} operate directly on the point cloud, but most \cite{qi2017pointnet++,hu2020randla,thomas2019kpconv} rely on downsampling to accommodate limited computational resources. 
KPConv \cite{thomas2019kpconv} introduces customized spatial kernel-based point convolution which yields the most favorable outcomes compared to other point-based methods. However, it require a significant amount of computational resources, which can limit its practicality in certain applications.

Projection-based methods \cite{xu2020squeezesegv3,cortinhal2020salsanext,zhang2020polarnet} involve projecting a 3D point cloud onto a 2D image plane, which can be done using either in spherical Range-View \cite{xu2020squeezesegv3,cortinhal2020salsanext}, Bird-Eye-View \cite{zhang2020polarnet}, or multi-view representations \cite{gerdzhev2021tornadonet}. 
These approaches usually operate in real-time, benefiting from efficient 2D CNN inference on GPUs. However, they suffer from significant information loss due to projection, which can prevent them from being on par with state-of-the-art methods.

The emergence of 3D voxel-based approaches \cite{cheng20212,zhu2021cylindrical,hou2022pvkd} can be attributed to recent advancements in sparse 3D convolutions \cite{choy20194d,tang2022torchsparse}. Typically, LiDAR point clouds are converted into Cartesian \cite{cheng20212} or Cylindrical \cite{zhu2021cylindrical} voxels and then processed using sparse convolutions. They have demonstrated state-of-the-art performance, but reducing voxel resolution can result in significant loss of information.

Recently, researchers have explored the potential of leveraging multiple data representations \cite{tang2020searching,yan20222dpass,xu2021rpvnet,ye2021drinet} to benefit from the information acquired by each representation. For instance, SPVNAS \cite{tang2020searching} incorporates both point-wise features and 3D-voxel features in their network while learning to segment. Similarly, RPVNet \cite{xu2021rpvnet} utilizes a trilateral structure in their network, where each branch adopts a distinct data representation mentioned earlier. 

Our method takes a similar approach to incorporate both point and voxel representations in the network. Voxel representations are utilized for contextual information learning, while point representations are employed to preserve fine-grained details, such as object boundaries and curvatures.

\subsection{Temporal LiDAR Reasoning}

Previous methods \cite{shi2020spsequencenet,schuett2022abstractflow,wang2022meta} have attempted to integrate a small number of past frames to facilitate temporal reasoning. For instance, SpSequenceNet \cite{shi2020spsequencenet} introduces cross-frame global attention, which uses information from the previous frame to emphasize features in the current frame. Additionally, it utilizes cross-frame local interpolation to combine local features from both the previous and current frame. Further, MetaRangeSeg \cite{wang2022meta} includes information from past frames by having the residual depth map as an additional feature to the input. However, these methods are not well-suited for continuous application to a temporal sequence of LiDAR point clouds. Firstly, they are inefficient as they discard previous observations and require the model to start from scratch each time a new point cloud is received. Secondly, they only aggregate temporal information within a short horizon, failing to capture the potential of temporal reasoning over a longer duration.

Several works \cite{duerr2020lidar,frossard2021strobe} have attempted to incorporate temporal reasoning through a recurrent framework. Specifically, Duerr \etal \cite{duerr2020lidar} presents a recurrent segmentation architecture in RV, which uses previous observations as memory and aligns them temporally in the range image space. However, RV memory has limitations, such as prioritizing nearby objects, and suffering from contention of memory resources during occlusions (since information from behind the occluder and the occluder itself are now sharing the same memory entries).
Further, StrObe \cite{frossard2021strobe} implemented a multi-scale 2D BEV memory for object detection enabling the network to take advantage of previous computations and process new LiDAR data incrementally. However, compared to object detection, semantic segmentation requires a more fine-grained and nuanced understanding of the scene and therefore a 3D memory is more suitable than 2D BEV. 

Our approach is distinct from previous methods as we propose an online LiDAR semantic segmentation framework utilizing a sparse 3D memory. By representing the memory in 3D, we preserve the distance between points regardless of viewpoint or distance from the SDV, allowing for size priors of different classes to be learned. Additionally, the memory also retains previously observed regions even when currently occluded, as occluders and occluded objects occupy different 3D regions, despite being in the same space in RV. Furthermore, we preserve fine-grained height details that are lost in the BEV memory.

%% file: sections/method.tex
\section{3D Memory-based LiDAR Segmentation}
\label{sec:method}

In this section we introduce \proposed, an online semantic segmentation framework for streaming LiDAR point clouds that leverages a 3D latent memory to remember the past and better handle occlusions and sparse observations. 
In the remainder of this section, we first describe our model formulation, then present the network architecture
and finally explain the learning process. 

\begin{figure*}[t]
    \centering
    \includegraphics[width=0.95\linewidth]{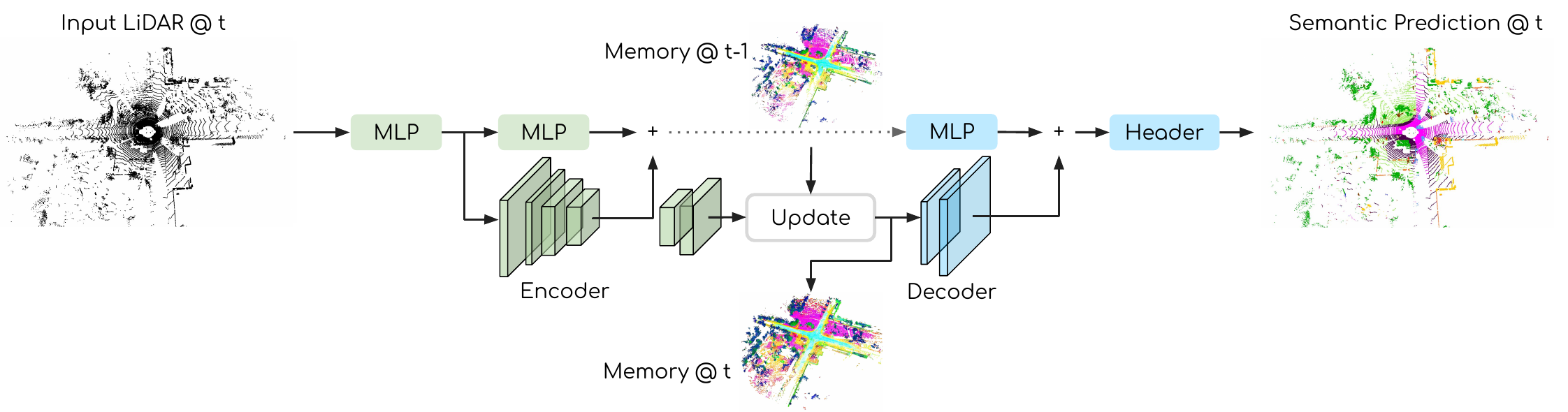}
    \caption{
        An overview of our model, \proposed.
        After the {\color{YellowGreen} encoder} processes the LiDAR point cloud at time $t$, the resulting feature map is utilized to update the latent memory (see Fig.~\ref{fig:update} for more details about the memory update). 
        Then, the {\color{Cerulean} decoder} combines the refined memory with the point embeddings from the encoder to obtain semantic predictions.
        }
    \label{fig:model}
\end{figure*}

\subsection{Model Formulation}

Let $\mathcal{P} = \{\mathcal{P}_t\}_{t=1}^{L}$ be a sequence of LiDAR sweeps where $L \in \mathbb{N}^{+}$ is the sequence length 
and $t \in [1, L]$ is the time index. 
Each LiDAR sweep $\mathcal{P}_t = (\mathbf{G}_t, \mathbf{F}_t)$ is a $\ang{360}$ scan of the surrounding with $N_t$ unordered points. $\mathbf{G}_t \in \mathbb{R}^{N_t \times 3}$ contains the Cartesian coordinates in the ego vehicle frame and $\mathbf{F}_t \in \mathbb{R}^{N_t}$ is the LiDAR intensity of the point. 
Let $\mathbf{T}_{t-1 \rightarrow t} \in SE(3)$ be   the pose transformation from the vehicle frame at time $t-1$ to $t$.

To make informed semantic predictions, in this paper we maintain a latent (or hidden) memory in 3D. 
This memory is sparse in nature since the majority of the 3D space is unoccupied. 
To represent this sparsity, we parametrize the memory at time $t$ using a sparse set of voxels containing the coordinates $H_{G,t} \in \mathbb{R}^{M_t \times 3}$ and their corresponding learned embeddings $H_{F,t} \in \mathbb{R}^{M_t \times d_m}$. $M_t$ is the number of voxel entries in the latent memory at time $t$ and $d_m$ is the embedding dimension.
Preserving the voxel coordinates is important to perform alignment as the reference changes when the SDV moves.
We utilize a voxel-based sparse representation as it provides notable computational benefits with respect to dense tensors as well as sparse representations at the point level without sacrificing performance.

Inference follows a three-step process that is repeated every time a new LiDAR sweep is available: (i) the encoder takes in the most recent LiDAR point cloud at current time $t$ and extracts point-level and voxel-level  observation embeddings, (ii) the latent memory is updated taking into account the voxel-level embeddings from the new observations, and (iii) the semantic predictions are decoded by combining the point-level 
embeddings from the encoder and voxel-level embeddings from the updated memory.
We refer the reader to Fig. \ref{fig:model} for an illustration of our method.

The memory update stage faces challenges due to the changing reference frame as the SDV moves, different sparsity levels of memory and current LiDAR sweep, as well as the motion of other actors. To address these challenges, a Feature Alignment Module (FAM) is introduced to align the previous memory state with the current observation embeddings. Subsequently, an Adaptive Padding Module (APM) is utilized to fill in missing observations in the current data and add new observations to the memory. Then, a Memory Refinement Module (MRM) is employed to update the latent memory using padded observations. Next, we explain each component in more detail.

\paragraph{Encoder:} 
Following  \cite{tang2020searching}, our encoder is composed of a point-branch that computes point-level embeddings preserving the fine details and a voxel-branch 
that performs contextual reasoning through 3D sparse convolutional blocks \cite{tang2022torchsparse}. 
The point-branch receives a 7-dimensional feature vector per point, with the xyz coordinates, intensity, and relative offsets to the nearest voxel center as features. It comprises two shared MLPs that output point embeddings, as illustrated in Fig.~\ref{fig:model}.
We average point embeddings belonging to the same voxel from the first shared MLP 
over voxels of size $v_b$ to obtain voxel features. These features are then processed through four residual blocks with 3D sparse convolutions, each downsampling the feature map by a factor of 2. 
Two additional residual blocks with 3D sparse convolutions are applied to upsample the sparse feature maps. Unlike a full U-Net \cite{ronneberger2015u} that recovers features at the original resolution, we only upsample to $\frac{1}{4}$ of the original size for  computational efficiency reasons, and use  the coarser features to update the latent memory before decoding finer details to output our semantic predictions.

\begin{figure}[t]
    \centering
    \includegraphics[width=\columnwidth]{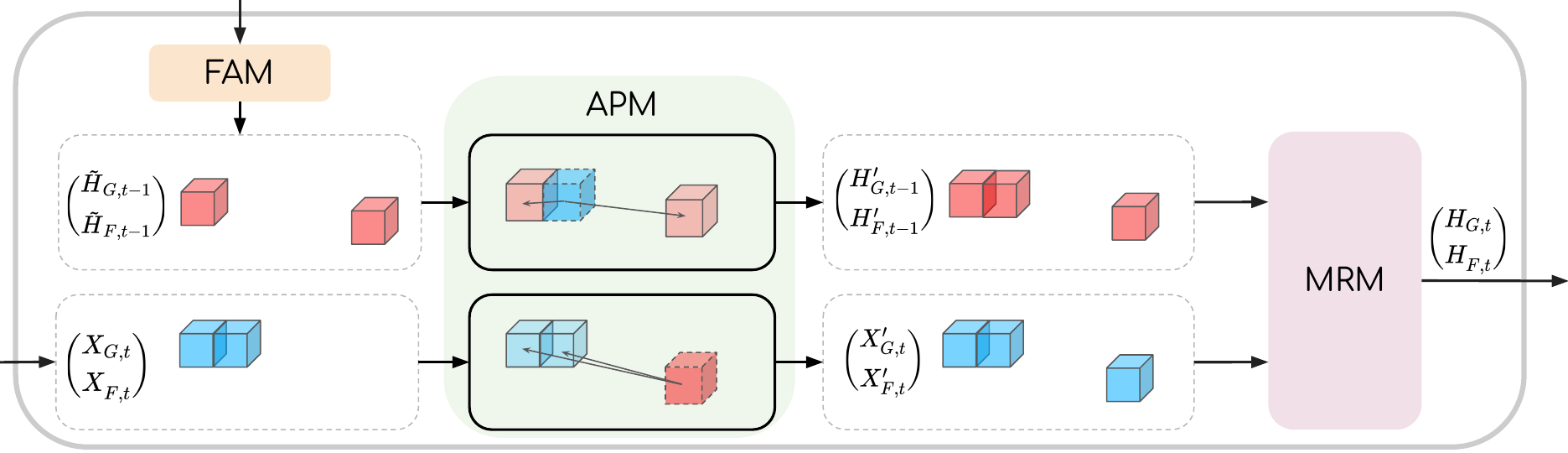}
    \caption{
    Overview of the latent memory update process.    
    The latent memory embeddings ($H_{G,t-1},H_{F,t-1}$) are transformed to the ego frame at $t$ with the Feature Alignment Module (FAM). Next, the Adaptive Padding Module (APM) is utilized to learn the padding of both memory and observation embeddings. The Memory Refinement Module (MRM) updates the latent memory by incorporating the padded observation embeddings. The updated memory is then passed to the decoder for generating semantic predictions. (Best viewed in color and zoomed-in.)
        }
    \label{fig:update}
    \vspace{-10px}
\end{figure}
\paragraph{Feature Alignment:}
The reference frame changes as the SDV moves.
We propose the Feature Alignment Module (FAM) to 
transform the latent memory from the ego frame at $t-1$ to $t$ and align with the current observation embeddings.
Specifically, we take the memory voxel coordinates $H_{G,t-1}$ and use the pose information $\mathbf{T}_{t-1 \rightarrow t}$ to project from the ego frame at $t-1$ to $t$. 
We then re-voxelize using the projected coordinates with a voxel size of $v_m$. 
If multiple entries are inside the same memory voxel, we take the average of them to be the voxel feature. The resulting warped coordinates and embeddings of the memory in the ego frame $t$ are denoted as $\tilde H_{G,t}$ and $\tilde H_{F,t}$, respectively. 

\paragraph{Adaptive Padding:}
To handle the different sparsity of the latent memory and the voxel-level observation embeddings,
we propose the Adaptive Padding Module (APM). 
We refer the readers to Fig.~\ref{fig:update} for an illustration.
First, we re-voxelize the encoder features with the same voxel size $v_m$ where entries within the same voxel are averaged. We denote the resulting coordinates and embeddings as $X_{G}$ and $X_{F}$. Note that we omit $t$ in this section for brevity.
Let $x_{G} \subseteq X_{G}$ and $x_{F} \subseteq X_{F}$ be the coordinates and embeddings 
of the new observations at time $t$ that are not present in the memory. 
To obtain an initial guess of the memory embedding for a new entry, we utilize a weighted aggregation approach within its surrounding neighbourhood. This involves taking into account the coordinate offsets relative to the existing neighbouring voxels in the memory, which provides insight into their importance for aggregation, similar to Continuous Conv \cite{wang2018deep}. In addition to this, we incorporate the feature similarities and feature distances as additional cues for the aggregation process. Encoding feature similarities is particularly useful for assigning weights to the neighborhood. In a dynamic scene with moving actors, the closest voxel may not always be the most important voxel. By providing feature similarities, the network can make more informed decisions.
The goal of such completion is to make a hypothesis of the embedding at the previously unobserved location using the available information. 
More precisely, we add those entries in the memory where their coordinates are $h_{G}' = x_{G}$ and the embedding of each voxel $j$ are initialized as follows,
\begin{equation}
    h_{F, j}' = \sum_{i \in \Omega_{\tilde H}(j)} w_{ji} \tilde H_{F, i},
\end{equation}
\begin{equation}
    w_{ji} = \psi(
        \tilde H_{G, i} - x_{G, j}, 
        \lVert \tilde H_{F, i} - x_{F, j} \rVert, 
        \frac{ \tilde H_{F, i}  \cdot x_{F, j}}{\lVert \tilde H_{F, i} \rVert \lVert x_{F, j} \rVert}
        ),
\end{equation}
where $i$ and $j$ are voxel indices, $\Omega_{\tilde H}(j)$ is the k-nearest-neighbourhood of voxel $j$ in $\tilde H_{G}$, and $\psi$ is a shared MLP followed by a softmax layer on the dimension of neighbourhood to ensure $\sum_{i} w_{ji} = 1$. 

Second, we identify the regions in the memory that are unseen in the current observation and denote their coordinates and embeddings as $\tilde h_{G} \subseteq \tilde H_{G}$ and $\tilde h_{F} \subseteq \tilde H_{F}$. We add entries $x_{G}'$ and $x_{F}'$ to complete the current observation in a similar manner.

\paragraph{Memory Refinement:} 
\label{sec:rum}

We design a sparse version of a ConvGRU \cite{ballas2016convgru} to update the latent memory $H'_{F, t-1}$ using the current padded observation embeddings $X'_{F, t}$ as follows:
\begin{equation}
    \begin{aligned}
        r_t = \text{sigmoid}[\Psi_r( X'_{F,t}, H'_{F, t-1})], \\ 
        z_t = \text{sigmoid}[\Psi_z( X'_{F,t}, H'_{F, t-1})], \\
        \hat{H}_{F,t} = \text{tanh}[\Psi_u( X'_{F,t}, r_t \cdot H'_{F, t-1})], \\
        H_{F,t} = \hat{H}_{F,t} \cdot z_t + H'_{F, t-1} \cdot(1 - z_t), \\
    \end{aligned}
\end{equation}
where $\Psi_r, \Psi_z, \Psi_u$ are sparse 3D convolutional blocks with downsampling layers that aim to expand the receptive field and upsampling layers that restore the embeddings to their original size. $r_t$ and $z_t$ are learned signals to reset or update the memory, respectively. We refer the readers to the supplementary material about the detailed architecture of the sparse convolutional blocks. 

\paragraph{Decoder:} 
Our decoder consists of an MLP, two residual blocks with sparse 3D convolutions, and a linear semantic header.
Specifically, we first take the corresponding memory embeddings at coordinates $\textbf{G}_t$ and add with the point embeddings from the encoder. The resulting combined embeddings are then voxelized with  voxel size of $\frac{v_b}{4}$ and further processed by two residual blocks that upsample the feature maps back to the original resolution. In parallel, an MLP takes the point embeddings before voxelization to retain the fine-grained details. Finally, the semantic header takes the combination of voxel and point embeddings to obtain per-point semantic predictions. 

\paragraph{Memory Initialization:}
At the start of the sequence ($t = 0$), the first observation is used to initialize memory, with $H_{G,0} = X_{G,0}$ and $H_{F,0} = X_{F,0}$. 

\subsection{Learning}
We learn our segmentation model by minimizing a linear combination of conventional segmentation loss functions \cite{hu2020randla,zhu2021cylindrical,tang2020searching} and a novel point-wise regularizer to better supervise the network training.
\begin{equation}
J =  \beta_{wce} J_{wce} +  \beta_{ls}  J_{ls} + \beta_{reg} J_{reg}.
\end{equation}
Here, $J_{wce}$ denotes cross-entropy loss, weighted by the inverse frequency of classes, to address class imbalance in the dataset.
The Lovasz Softmax Loss ($J_{ls}$) \cite{berman2018lovasz} is used to train the network, as it is a differentiable surrogate for the non-convex intersection over union (IoU) metric, which is a commonly used evaluation metric for semantic segmentation. Additionally, $J_{reg}$ corresponds to our proposed point-wise regularizer.
$\beta_{reg}$, $\beta_{wce}$ and $\beta_{ls}$ are hyperparameters.

\paragraph{Point-wise Smoothness:} Our regularizer is designed to limit significant variations in semantic predictions within the 3D neighborhood of each point, except when these variations occur at the class boundary.
Formally,
\begin{equation}
    \begin{aligned}
        J_{reg} = \frac{1}{N_t} \sum_{i=1}^{N_t} \lvert \Delta(Y, i) - \Delta(\hat Y, i) \rvert, \\
        \Delta(Y, i) = \Bigl| \frac{1}{\lvert \Omega_{\mathcal{P}_t}(i) \rvert} \sum_{j \in \Omega_{\mathcal{P}_{t}}(i)} \lvert y_i - y_j \rvert \Bigr|. \\
    \end{aligned}
\end{equation}
Here, $\Delta(Y, i)$ represents the ground truth semantic variation around point $i$, while $\Delta(\hat Y, i)$ corresponds to the predicted semantic variation around point $i$.
We use $\hat Y \in \mathbb{R}^{N_t \times C}$ to denote the predicted semantic distribution over $C$ classes, and $Y \in \mathbb{R}^{N_t \times C}$ to denote the ground truth semantic one-hot label. The variable $y_i$ represents the $i$-th element of $Y$.
$\Omega_{\mathcal{P}_{t}}(i)$ denotes the neighborhood of point $i$ in $\mathcal{P}_t$, and $\lvert \Omega_{\mathcal{P}_t}(i) \rvert$ represents the number of points in the neighborhood.
The inspiration for this regularizer comes from a study by \cite{gerdzhev2021tornadonet} that penalizes adjacent pixel prediction variations with the ground truth.

%% file: sections/experiments.tex
\section{Experiments}
In this section, we thoroughly analyze the performance of our method on three different datasets: SemanticKITTI \cite{behley2019semantickitti}, nuScenes \cite{caesar2020nuscenes}, and PandaSet \cite{xiao2021pandaset}.
These datasets were collected using different LiDAR sensors and in diverse geographical regions.
Our results demonstrate that \proposed{} outperforms the current state of the art in all benchmarks. Additionally, we conduct ablation studies to understand the impact of our contributions. 
We demonstrate that incorporating a 3D sparse latent memory improves semantic predictions, with a very significant gain in long-range areas as these are sparser and more often partially obstructed than short-range regions.

\input{figs/sk_multi_test_table}

\paragraph{Datasets:} 
We benchmark our approach on three large-scale autonomous driving datasets:
\textbf{SemanticKITTI} \cite{behley2019semantickitti} consists of 22 sequences  from the KITTI Odometry Dataset \cite{geiger2012kitti}, which was collected in Germany using a Velodyne HDL-64E LiDAR. We use the standard split where sequences 0 to 10 are used for training (with sequence 8 for validation), and sequences 11 to 21 are held for testing. 
We follow the standard setting where semantic labels are mapped to 19 classes for the single-scan benchmark and 25 for the multi-scan benchmark. 
In the latter, all movable classes are further divided into moving and non-moving classes. 
Given that prior research on temporal aggregation for semantic segmentation is more prevalent in the multi-scan benchmark, we consider it more appropriate for evaluating our proposed method. 
Single-scan benchmark results are also provided in the supplementary material.
\textbf{nuScenes} \cite{caesar2020nuscenes} was collected in Boston and Singapore using a Velodyne HDL-32E LiDAR sensor, which captures  sparser point clouds that pose extra challenges for semantic reasoning. The dataset contains a total of 1000 scenes, with 700 scenes used for training, 150 for validation, and the remaining 150 held out for testing. Each scene contains about 40 labeled frames recorded at 2 Hz. We follow the standard setting which maps 31 fine-grained classes to 16 classes for training and evaluation. 
\textbf{PandaSet} \cite{xiao2021pandaset} provides short sequences of temporal data but with a large number of moving actors in Silicon Valley. 
This dataset was collected using a Hesai-Pandar64 LiDAR and consists of 103 sequences, each with a length of 8 seconds. Following \cite{duerr2020lidar}, we grouped the original fine-grained semantic classes into 14 for training and testing, trying to match the classes in other popular segmentation benchmarks \cite{behley2019semantickitti,caesar2020nuscenes}.
We also used the same train/validation/test split as in the reference paper.

\paragraph{Implementation details:}
For all  datasets, we first train the network without the memory update for single-scan segmentation for 50 epochs. Then, we freeze the encoder and train the memory update block and decoder in a recurrent fashion for an additional 20 epochs.  During each training iteration, we use the first 10 consecutive sweeps as warmup memory, and then train the network with backpropagation through time (BPTT) on the next 3 frames. 
The memory voxel size $v_m$ is 0.5 m, and the embedding dimension $d_m$ is 128. We use the AdamW optimizer with a starting learning rate of $0.003$ and a decay factor of $0.9$. During training, we use  data augmentation including global scaling sampled at random from [0.80, 1.20], translation sampled from [0, 0.2] m on all three axes, and global rotation around the Z axis with a random angle sampled from $[-\pi,\pi]$. 
We set the loss weights $\beta_{wce}$ to 1, $\beta_{ls}$ to 2, and $\beta_{reg}$ to 500.
We set the regularizer neighborhood $\Omega_{P_t}(i)$ to be the closest 32 points around point $i$, and use a padding neighbourhood $\Omega_{\tilde{H}}(j)$ to be the closest 5 voxels of voxel $j$. All experiments are conducted using 4 NVIDIA T4 GPUs with a batch size of 1 per GPU.
During evaluation, we unroll each sequence from the first frame to the end and follow \cite{zhu2021cylindrical,hou2022pvkd} to apply test-time-augmentation, \ie averaging prediction scores of 10 forward passes with augmented input from a single model. 

To adapt to the uniqueness of each dataset, we consider the following per-dataset settings.
To train on the SemanticKITTI dataset, we set the voxel size $v_b$ to 0.05 m. Due to the limited quantities of movable actors in this dataset, following \cite{zhou2021panoptic} we generate a library of instances from the training sequences  and randomly inject 5 objects over the ground classes at each frame during training. 
To improve the ability to separate classes into moving and non-moving actors, we add another linear header to classify points as moving or static and fuse the results from the semantic header. More implementation details are provided in the supplementary material.
For nuScenes, the voxel size is set to 0.1 m, and 0.125 m for PandaSet.
\input{figs/nusc_test_table}
\input{figs/pd_test_table_full}
\paragraph{Metrics:}
We follow existing works \cite{zhu2021cylindrical, cheng20212,xu2021rpvnet} and use the intersection over union (IoU) averaged over all classes as the main metric (mIoU). Class-wise IoUs are also reported. 

\paragraph{Comparison against state-of-the-art:} 
Our results are compared with other state-of-the-art approaches on the test set of three datasets in Tab.~\ref{tab:skitti_multi} to \ref{tab:pandaset}. We include approaches that use LiDAR only for fair comparison.
Notably, our proposed method outperforms all prior works on all three datasets, demonstrating its generalizability across various LiDAR sensors and different geographical regions.

Tab.~\ref{tab:skitti_multi} presents a comparison of our approach on the SemanticKITTI multi-scan benchmark, which is designed to evaluate methods that focus on temporal aggregation for semantic segmentation. The other methods in the table are divided into two groups. 
The first group consists of baselines that are designed for single-scan segmentation but aggregate multiple consecutive frames as input at the cost of higher memory consumption and longer runtime.
In contrast, the second group proposes temporal aggregation modules specifically to handle multiple scans. It is worth noting that our proposed method performs significantly better than all of these approaches.
\proposed{} demonstrates exceptional performance in segmenting dynamic actors, such as \textit{moving bicyclist}, \textit{motorcyclist}, and \textit{person}. These substantial improvements can be attributed to the proposed recurrent framework, which implicitly encode the motion of moving actors in the 3D latent memory. 
Furthermore, we also demonstrate the effectiveness of our approach on the single-scan benchmark of the same dataset. We refer the readers to the supplementary material for the detailed results. This is a more competitive benchmark focusing on single-scan semantic segmentation where previous research has focused on proposing various architectures \cite{cheng20212,xu2021rpvnet,zhu2021cylindrical} or knowledge distillation techniques \cite{hou2022pvkd}. Our results show that \proposed{} can outperform these methods, which are highly optimized for this benchmark.

Results on nuScenes \cite{caesar2020nuscenes} benchmark are presented in Tab.~\ref{tab:nusc}. \proposed{} outperforms the previous state-of-the-art. Our method is particularly effective in smaller classes such as \textit{barriers}, \textit{pedestrians}, and \textit{traffic cones}, which present challenges for semantic segmentation networks due to the sparser point clouds available in this dataset. However, our approach overcomes this limitation by using a 3D latent memory to improve semantic reasoning of the sparse points. Additionally, we observe significant gains in the background classes, such as \textit{sidewalk}, \textit{terrain}, and \textit{manmade}, which often rely on understanding of the surrounding to be segmented correctly. Our method improves the contextual reasoning by accumulating past observations using a latent memory representation.

Finally, we compare our results with the state-of-the-art methods on PandaSet \cite{xiao2021pandaset} and report the test set IoUs in Tab.~\ref{tab:pandaset}. Our proposed method also achieves state-of-the-art performance in this dataset. Notably, our approach outperforms others by a large margin in almost all classes. We highlight that \proposed{} consistently outperforms TemporalLidarSeg \cite{duerr2020lidar} across multiple benchmarks, a method that also proposes a latent memory but in RV instead of 3D. These findings confirm our hypothesis that the 3D latent memory approach is more effective than using RV in accurately segmenting objects in 3D point clouds.

\begin{figure}[t]
    \centering
    \includegraphics[width=\linewidth]{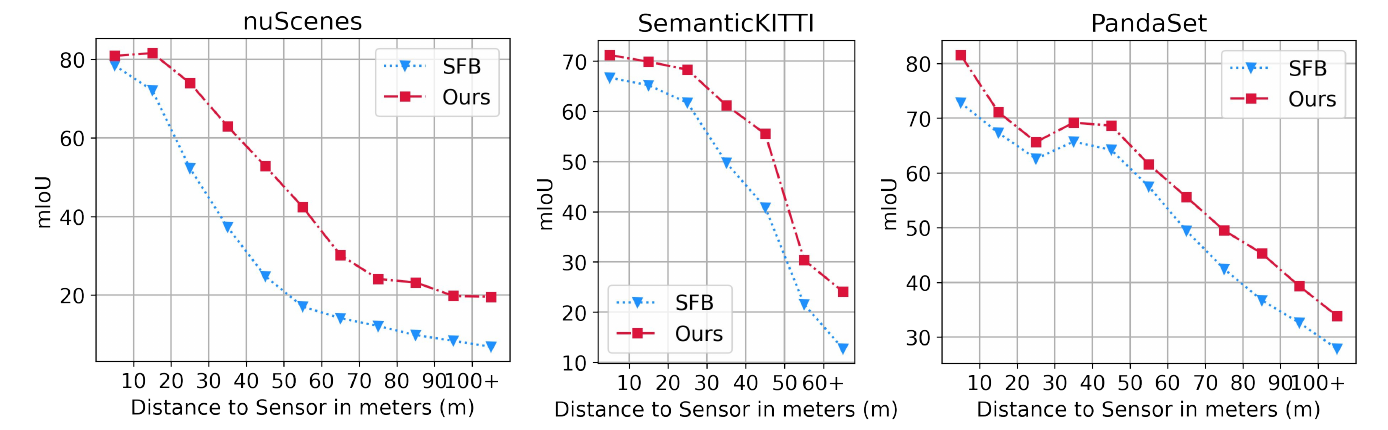}
    \caption{
        Comparison of \proposed{} with single frame baseline (SFB) on the validation set with different distance-range. 
        }
    \label{fig:range}
    \vspace{-10px}
\end{figure}

\begin{figure}[t]
    \centering
    \includegraphics[width=\linewidth]{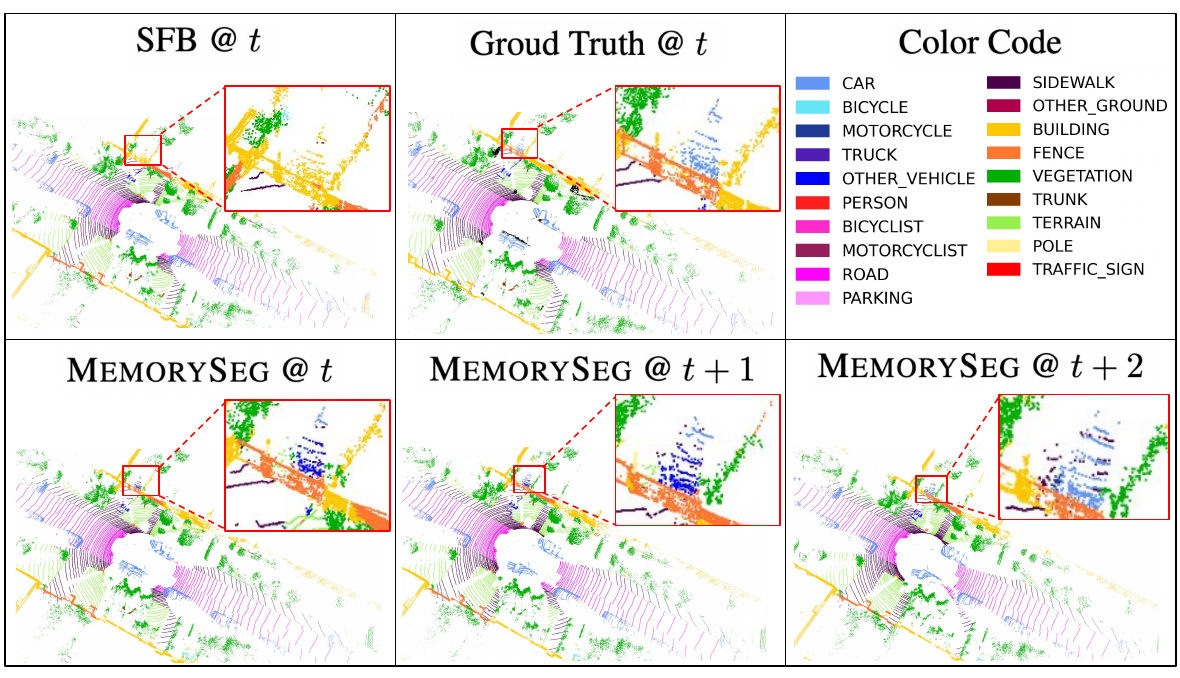}
    \caption{
        Predictions of \proposed{} over time are illustrated in the bottom row. We include the prediction from the single frame baseline (SFB) and ground truth with color code at the top.
    }
    \label{fig:qualitative}
\end{figure}

\paragraph{Importance of the memory: }
Fig.~\ref{fig:range} highlights the significance of incorporating memory, particularly in longer range regions where point clouds are much sparser. To assess the effectiveness of the learned latent memory, we compare against our encoder and decoder without the memory update module as the single frame baseline (SFB). Our proposed network consistently outperforms the SFB in all regions, with the most significant improvement observed in the long-range region of nuScenes, where points are sparser and more difficult to contextualize. Thanks to the 3D sparse latent memory that accumulates semantic embeddings of past frames, our proposed approach performs much better than the SFB, particularly in those challenging regions.
Fig.~\ref{fig:qualitative} provides similar insights. In the top-left of the figure, the SFB fails to segment a partially occluded vehicle located behind a fence. In contrast, our proposed approach initially segments the vehicle as \textit{other vehicle} class, but as the ego vehicle advances, the 3D latent memory accumulates and refines through new observations, enabling the network to correctly identify the vehicle as a \textit{car} after two sweeps. This illustration further confirms that the proposed 3D latent memory provides better contextualization of sparse objects that are partially occluded. Additional qualitative results are provided in the supplementary materials.
We also quantitatively demonstrate the importance of memory in Tab.~\ref{tab:componentsmall}. M1 builds on SFB and incorporates FAM and MRM. 
Note that both FAM and MRM are essential for updating the latent memory continuously. FAM aligns the sparse 3D latent memory in the traveling ego frame, while MRM helps in learning to reset and update the memory recurrently. APM is replaced with zero-padding in M1 to highlight the significance of memory.
M1 shows a $2.3\%$ improvement in the mIoU metric compared to SFB, attributed to the ability of the sparse 3D latent memory to aggregate information in the temporal dimension as the ego moves through the scene.


\paragraph{Influence of adaptive padding:}
In this section, we compare our proposed APM with zero padding and Continuous Conv \cite{wang2018deep}. The results are presented in Tab.~\ref{tab:componentsmall}.
M2 is an extension of M1 that introduces adaptive padding on both observation and memory embeddings when they are aligned. However, the padded entries are only processed using the relative coordinates in the neighborhood, similar to Continuous Conv \cite{wang2018deep} with the attention mechanism.
A slight improvement of $0.2 \%$ is observed when new observations in the latent memory are initialized with a learned guess instead of all zeros.
To further improve the performance of the network, M4 exploits feature similarities and distances in APM. When assigning an initial embedding to a new entry, both the relative position and the similarity to neighbors are considered. This is particularly useful for moving actors in the scene, where the network should learn to start with a guess with a similar embedding in the neighborhood, rather than the closest entry.
By comparing M0 with M3 further confirms the usefulness of the proposed APM.
\paragraph{Influence of memory refinement:}
In M3 of Tab.~\ref{tab:componentsmall}, we replace our proposed MRM with a vanilla sparse ConvGRU \cite{ballas2016convgru}, which has a limited receptive field due to the convolution kernel size, making it challenging for fast-moving actors. However, MRM overcomes this limitation by downsampling features to increase the receptive field. This improves the learning of meaningful reset and update gates for refining the latent memory.

\input{figs/ablation_components_table}
\input{figs/ablation_reg_table}
\paragraph{Influence of the regularizer:} 
To demonstrate the effectiveness of our regularizer to train semantic segmentation networks on point clouds with different characteristics, we ablate this loss on the three datasets and summarize the results in Tab.~\ref{tab:reg}. The results consistently show that incorporating the regularizer leads to performance improvements. The regularizer provides additional supervision on variations and boundaries, which is particularly beneficial for nuScenes. We note once again that this dataset has sparser point clouds than the other two.

\paragraph{Parameter count and runtime comparison:} 
\proposed \ incurs only a relatively small computational overhead of 23\% more parameters and 20\% runtime than SFB. On the other hand, when naively concatenating the past five frames as the input to the model, despite having the same number of parameters as SFB, it takes 1.58 times longer (32\% slower than our method) while only exploiting a limited 0.5-second window from the past.

%% file: figs/sk_multi_test_table.tex
\begin{table*}[ht]
    {
    \centering
    \resizebox{\textwidth}{!}{
    \begin{tabular}{l|c|ccccccccccccccccccccccccc}
    \hline 
    \textbf{Method}  
    & \begin{sideways} \textbf{mIoU} \end{sideways} 
    & \begin{sideways} Car \end{sideways} 
    & \begin{sideways} Bicycle \end{sideways} 
    & \begin{sideways} Motorcycle \end{sideways} 
    & \begin{sideways} Truck \end{sideways} 
    & \begin{sideways} Other Vehicle \end{sideways} 
    & \begin{sideways} Person \end{sideways} 
    & \begin{sideways} Bicyclist \end{sideways} 
    & \begin{sideways} Motorcyclist \end{sideways} 
    & \begin{sideways} Road \end{sideways} 
    & \begin{sideways} Parking \end{sideways} 
    & \begin{sideways} Sidewalk \end{sideways} 
    & \begin{sideways} Other Ground \end{sideways} 
    & \begin{sideways} Building \end{sideways} 
    & \begin{sideways} Fence \end{sideways} 
    & \begin{sideways} Vegetation \end{sideways} 
    & \begin{sideways} Trunk \end{sideways} 
    & \begin{sideways} Terrain \end{sideways} 
    & \begin{sideways} Pole \end{sideways} 
    & \begin{sideways} Traffic Sign \end{sideways} 
    & \begin{sideways} car (m) \end{sideways} 
    & \begin{sideways} bicyclist (m) \end{sideways} 
    & \begin{sideways} person (m) \end{sideways} 
    & \begin{sideways} motorcyclist (m) \end{sideways} 
    & \begin{sideways} other-vehicle (m) \end{sideways} 
    & \begin{sideways} truck (m) \end{sideways} \\ 
    \hline
    TangentConv \cite{tatarchenko2018tangent} & $34.1$ & $84.9$ & $2.0$ & $18.2$ & $21.1$ & $18.5$ & $1.6$ & $0.0$ & $0.0$ & $83.9$ & $38.3$ & $64.0$ & $15.3$ & $85.8$ & $49.1$ & $79.5$ & $43.2$ & $56.7$ & $36.4$ & $31.2$ & $40.3$ & $1.1$ & $ 6.4$ & $ 1.9$ & $\mathbf{30.1}$ & $\mathbf{42.2}$ \\
    DarkNet53Seg \cite{behley2019semantickitti} & $41.6$ & $84.1$ & $30.4$ & $32.9$ & $20.2$ & $20.7$ & $7.5$ & $0.0$ & $0.0$ & $91.6$ & $64.9$ & $75.3$ & $27.5$ & $85.2$ & $56.5$ & $78.4$ & $50.7$ & $64.8$ & $38.1$ & $53.3$ & $61.5$ & $14.1$ & $15.2$ & $0.2$ & $28.9$ & $37.8$ \\
    KPConv \cite{thomas2019kpconv} & $51.2$ & $93.7$ & $44.9$ & $47.2$ & $42.5$ & $38.6$ & $21.6$ & $0.0$ & $0.0$ & $86.5$ & $58.4$ & $70.5$ & $26.7$ & $90.8$ & $64.5$ & $84.6$ & $70.3$ & $66.0$ & $57.0$ & $53.9$ & $69.4$ & $67.4$ & $67.5$ & $47.2$ & $4.7$ & $5.8$ \\
    Cylinder3D \cite{zhu2021cylindrical} & $52.5$ & $\mathbf{94.6}$ & $67.6$ & $63.8$ & $41.3$ & $38.8$ & $12.5$ & $\mathbf{1.7}$ & $0.2$ & $90.7$ & $\mathbf{65.0}$ & $74.5$ & $\mathbf{32.3}$ & $\mathbf{92.6}$ & $66.0$ & $\mathbf{85.8}$ & $72.0$ & $68.9$ & $63.1$ & $61.4$ & $\mathbf{74.9}$ & $68.3$ & $65.7$ & $11.9$ & $0.1$ & $0.0$ \\
    \hline
    SpSequenceNet \cite{shi2020spsequencenet} & $43.1$ & $88.5$ & $24.0$ & $26.2$ & $29.2$ & $22.7$ & $6.3$ & $0.0$ & $0.0$ & $90.1$ & $57.6$ & $73.9$ & $27.1$ & $91.2$ & $66.8$ & $84.0$ & $66.0$ & $65.7$ & $50.8$ & $48.7$ & $53.2$ & $41.2$ & $26.2$ & $36.2$ & $2.3$ & $ 0.1$ \\
    TemporalLidarSeg \cite{duerr2020lidar} & $47.0$ & $92.1$ & $47.7$ & $40.9$ & $39.2$ & $35.0$ & $14.4$ & $0.0$ & $0.0$ & $\mathbf{91.8}$ & $59.6$ & $\mathbf{75.8}$ & $23.2$ & $89.8$ & $63.8$ & $82.3$ & $62.5$ & $64.7$ & $52.6$ & $60.4$ & $68.2$ & $42.8$ & $40.4$ & $12.9$ & $12.4$ & $2.1$ \\
    TemporalLatticeNet \cite{schuett2022abstractflow} & 47.1 & 91.6 & 35.4 & 36.1 & 26.9 & 23.0 & 9.4 & 0.0 & 0.0 & 91.5 & 59.3 & 75.3 & 27.5 & 89.6 & 65.3 & 84.6 & 66.7 & $\mathbf{70.4}$ & 57.2 & 60.4 & 59.7 & 41.7 & 51.0 & 48.8 & 5.9 & 0.0\\
    Meta-RangeSeg \cite{wang2022meta} & 49.7 & 90.8 & 50.0 & 49.5 & 29.5 & 34.8 & 16.6 & 0.0 & 0.0 & 90.8 & 62.9 & 74.8 & 26.5 & 89.8 & 62.1 & 82.8 & 65.7 & 66.5 & 56.2 & 64.5 & 69.0 & 60.4 & 57.9 & 22.0 & 16.6 & 2.6 \\
    \hline
    \proposed \ [ours] & $\mathbf{58.3}$ & 94.0 & $\mathbf{68.3}$ & $\mathbf{68.8}$ & $\mathbf{51.3}$ & $\mathbf{40.9}$ & $\mathbf{27.0}$ & 0.3 & $\mathbf{2.8}$ & 89.9 & 64.3 & 74.8 & 29.2 & 92.2 & $\mathbf{69.3}$ & 84.8 & $\mathbf{75.1}$ & 70.1 & $\mathbf{65.5}$ & $\mathbf{68.5}$ & 71.7 & $\mathbf{74.4}$ & $\mathbf{71.7}$ & $\mathbf{73.9}$ & 15.1 & 13.6 \\
    \hline
    \end{tabular}
    }
    \caption{Comparison to the state-of-the-art methods on the test set of SemanticKITTI \cite{behley2019semantickitti} multi-scan benchmark. (m) indicates moving. We include LiDAR-only published approaches at the time of submission. 
    These approaches are categorized into two groups: the first group includes methods that aim for single-scan segmentation but use multiple aggregated frames as input; the second group consists of methods that introduce temporal aggregation modules explicitly tailored for handling multiple scans.
    Metrics are provided in [\%].
    }
    \label{tab:skitti_multi} }
    \vspace{-10px}
    \end{table*} 

%% file: figs/nusc_test_table.tex
\begin{table*}[ht]
    {
    \centering
    \resizebox{\textwidth}{!}{
    \begin{tabular}{l|cc|cccccccccccccccc}
    \hline 
    \textbf{Method}  
    & \begin{sideways} \textbf{mIoU} \end{sideways} 
    & \begin{sideways} \textbf{FW-mIoU} \end{sideways} 
    & \begin{sideways} Barrier \end{sideways} 
    & \begin{sideways} Bicycle \end{sideways} 
    & \begin{sideways} Bus \end{sideways} 
    & \begin{sideways} Car \end{sideways} 
    & \begin{sideways} Construction \end{sideways} 
    & \begin{sideways} Motorcycle \end{sideways} 
    & \begin{sideways} Pedestrain \end{sideways} 
    & \begin{sideways} Traffic Cone \end{sideways} 
    & \begin{sideways} Trailer \end{sideways} 
    & \begin{sideways} Truck \end{sideways} 
    & \begin{sideways} Drivable \end{sideways} 
    & \begin{sideways} Other Flat \end{sideways} 
    & \begin{sideways} Sidewalk \end{sideways} 
    & \begin{sideways} Terrain \end{sideways} 
    & \begin{sideways} Manmade \end{sideways} 
    & \begin{sideways} Vegetation \end{sideways}  \\ 
    \hline
    PolarNet \cite{zhang2020polarnet} & $69.4$ & $ 87.4$ & $72.2$ & $16.8$ & $77.0$ & $86.5$ & $51.1$ & $69.7$ & $64.8$ & $54.1$ & $69.7$ & $63.5$ & $96.6$ & $67.1$ & $77.7$ & $72.1$ & $87.1$ & $84.5$ \\
    Cylinder3D \cite{zhu2021cylindrical} & $77.2$ & $ 89.9$ & $82.8$ & $ 29.8$ & $84.3$ & $89.4$ & $63.0$ & $\mathbf{79.3}$ & $77.2$ & $73.4$ & $84.6$ & $69.1$ & $97.7$ & $ \mathbf{70.2}$ & $ 80.3$ & $75.5$ & $90.4$ & $87.6$ \\
    SPVCNN \cite{tang2020searching} & $77.4$ & $ 89.7$ & $80.0$ & $ 30.0$ & $\mathbf{91.9}$ & $90.8$ & $64.7$ & $79.0$ & $75.6$ & $70.9$ & $81.0$ & $74.6$ & $97.4$ & $69.2$ & $80.0$ & $76.1$ & $89.3$ & $87.1$ \\
    (AF)2-S3Net \cite{cheng20212} & $78.3$ & $ 88.5$ & $78.9$ & $ \mathbf{52.2}$ & $89.9$ & $84.2$ & $ \mathbf{77.4}$ & $ 74.3$ & $77.3$ & $72.0$ & $83.9$ & $73.8$ & $97.1$ & $66.5$ & $77.5$ & $74.0$ & $87.7$ & $86.8$ \\
    \hline
    \proposed \ [ours] & $\mathbf{80.6}$ & $\mathbf{91.4}$ & $\mathbf{84.9}$ & 40.2 & 91.2 & $\mathbf{92.4}$ & 71.2 & 73.5 & $\mathbf{85.9}$ & $\mathbf{77.8}$ & $\mathbf{88.0}$ & $\mathbf{76.4}$ & $\mathbf{97.9}$ & 69.0 & $\mathbf{81.2}$ & $\mathbf{77.6}$ & $\mathbf{92.6}$ & $\mathbf{89.7}$ \\
    \hline
    \end{tabular}
    }
    \caption{Comparison to the state-of-the-art methods on the test set of nuScenes \cite{caesar2020nuscenes} LiDAR semantic segmentation benchmark. We include LiDAR-only published approaches at the time of submission.
    Metrics are provided in [\%]}
    \label{tab:nusc} }
    \end{table*} 

%% file: figs/pd_test_table_full.tex
    \begin{table*}[ht]
        {
        \centering
        \resizebox{0.8\textwidth}{!}{
        \begin{tabular}{l|c|cccccccccccccc}
        \hline 
        \textbf{Method}  
        & \begin{sideways} \textbf{mIoU} \end{sideways} 
        & \begin{sideways} Car \end{sideways} 
        & \begin{sideways} Bicycle \end{sideways} 
        & \begin{sideways} Motorcycle \end{sideways} 
        & \begin{sideways} Truck \end{sideways} 
        & \begin{sideways} Other Vehicle \end{sideways} 
        & \begin{sideways} Person \end{sideways} 
        & \begin{sideways} Road \end{sideways} 
        & \begin{sideways} Road Barriers \end{sideways} 
        & \begin{sideways} Sidewalk \end{sideways} 
        & \begin{sideways} Building \end{sideways} 
        & \begin{sideways} Vegetation \end{sideways} 
        & \begin{sideways} Terrain \end{sideways} 
        & \begin{sideways} Background \end{sideways} 
        & \begin{sideways} Traffic Sign \end{sideways} \\ 
        \hline
        SqueezeSegv3 \cite{xu2020squeezesegv3} & $55.7$ & $92.8$ & $24.1$ & $18.0$ & $36.5$ & $54.3$ & $63.0$ & $91.1$ & $11.9$ & $71.3$ & $86.2$ & $85.0$ & $61.3$ & $63.2$ & $20.6$ \\
        SalsaNext \cite{cortinhal2020salsanext} & $57.8$ & $92.1$ & $40.7$ & $31.7$ & $28.7$ & $56.2$ & $69.0$ & $90.0$ & $22.6$ & $67.1$ & $85.6$ & $83.4$ & $58.5$ & $63.3$ & $20.6$ \\
        TemporalLidarSeg \cite{duerr2020lidar} & $60.0$ & $93.7$ & $33.6$ & $38.0$ & $37.1$ & $59.9$ & $72.0$ & $91.1$ & $14.6$ & $70.6$ & $88.2$ & $88.4$ & $63.8$ & $68.4$ & $20.7$ \\
        SPVCNN \cite{tang2020searching} & 64.7 & 95.8 & 38.1 & 46.3 & 44.0 & 74.1 & 78.6 & 91.2 & $\mathbf{28.3}$ & 70.3 & 87.2 & 87.5 & 61.5 & 67.5 & 35.6 \\
        \hline
        \proposed \ [ours] & $\mathbf{70.3}$ & $\mathbf{97.2}$ & $\mathbf{60.2}$ & $\mathbf{58.4}$ & $\mathbf{62.9}$ & $\mathbf{74.3}$ & $\mathbf{82.6}$ & $\mathbf{92.1}$ & 27.7 & $\mathbf{74.1}$ & $\mathbf{89.4}$ & $\mathbf{90.7}$ & $\mathbf{64.9}$ & $\mathbf{72.8}$ & $\mathbf{36.4}$ \\
    
        \hline
        \end{tabular}
        }
        \caption{Comparison to the state-of-the-art methods on the test set of PandaSet \cite{xiao2021pandaset}. Metrics are provided in [\%]. 
        }
            \label{tab:pandaset}
        }
        \vspace{-10px}
        \end{table*} 

%% file: figs/ablation_components_table.tex
            \begin{table}[t]\centering
                \scriptsize
                \begin{tabular}{lrrrrrrrrcr}\toprule
                &\textbf{FAM} & &\multicolumn{2}{c}{\textbf{APM}} & &\multicolumn{2}{c}{\textbf{MRM}} & &\textbf{mIoU} \\\cmidrule{2-2}\cmidrule{4-5}\cmidrule{7-8}\cmidrule{10-10}
                & & &\textbf{CC} &\textbf{Ours} & &\textbf{SCG} &\textbf{Ours} & & \\\midrule
                SFB & & & & & & & & &67.2 \\
                M0 &\checkmark & &\checkmark & & &\checkmark & & &68.9 \\
                M1 &\checkmark & & & & & &\checkmark & &69.5 \\
                M2 &\checkmark & &\checkmark & & & &\checkmark & &69.7 \\
                M3 &\checkmark & & &\checkmark & &\checkmark & & &69.7 \\
                M4 &\checkmark & & &\checkmark & & &\checkmark & &$\mathbf{70.8}$ \\
                \bottomrule
                \end{tabular}
                \caption{Ablation on the proposed components in the network. Metrics are mIoU on the validation set of SemanticKITTI \cite{behley2019semantickitti} provided in [\%].
            \textbf{FAM}: Feature Alignment Module,
            \textbf{APM}: Adaptive Padding Module,
            \textbf{CC}: padding using Continuous Conv \cite{wang2018deep},
            \textbf{SCG}: Sparse ConvGRU \cite{ballas2016convgru}.
            \textbf{MRM}: Memory Refinement Module.
            }
                \label{tab:componentsmall}
                \end{table}

%% file: figs/ablation_reg_table.tex
\begin{table}[t]
    \begin{center}
    \scalebox{0.8}
    {
    \begin{tabular}{c|ccc}
    \hline 
    \textbf{Regularizer} & \textbf{SemanticKITTI} & \textbf{nuScenes} & \textbf{PandaSet} \\
    \hline \hline
    & $66.4$ & $72.9$ & $64.7$ \\
    \checkmark & $\mathbf{67.2}$ & $\mathbf{76.7}$ & $\mathbf{66.6}$ \\
    \hline
    \end{tabular}
    }
    \end{center}
    \caption{Ablation on the model performance with and without the proposed regularizer. Metrics are mIoU on the validation set provided in [\%].}
    \label{tab:reg}
    \vspace{-10px}
    \end{table}

%% file: sections/conclusion.tex
\section{Conclusion}
In this paper, we have proposed a novel online LiDAR segmentation model named \proposed, which utilizes a sparse 3D latent memory to recurrently accumulate learned semantic embeddings from past observations. We also presented a novel variation regularizer to supervise the learning of 3D semantic segmentation on point clouds. Our results demonstrate that our approach achieve state-of-the-art performance on three large-scale datasets. This improved performance can be attributed to the ability of the latent memory to better contextualize objects in the scene. Looking ahead, our future work will focus on integrating instance segmentation and tracking to develop an end-to-end memory-augmented panoptic segmentation framework. 

%% file: sections/supp_details.tex
\section{Implementation Details}
\input{sections/supp_mrm}
\subsection{Details on Motion Header}
\label{sec:motion}
\begin{figure}[t]
  \centering
  \includegraphics[width=\linewidth]{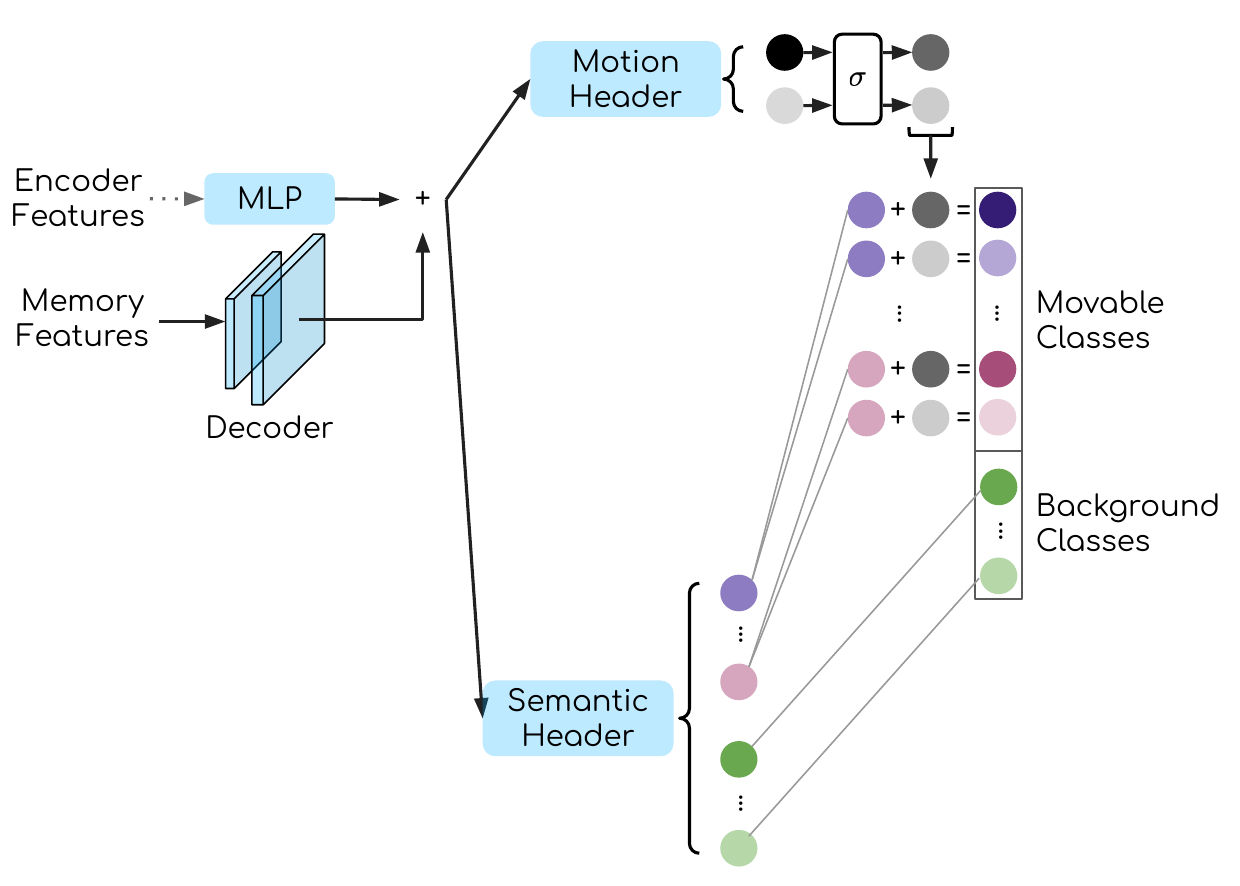}
  \caption{Illustration of the modification of the decoder for the multi-scan benchmark of SemanticKITTI \cite{behley2019semantickitti}. $\sigma$ denotes the \textit{LogSoftMax} operation to normalize the motion features. The motion logits are broadcasted and added to the movable logtis from the semantic header in the \textit{Log} space.}
    \label{fig:motion_header}
  \vspace{-10px}
\end{figure}

In this section, we explain how we implemented the motion header in the decoder to classify movable actors as either moving or non-moving in the SemanticKITTI dataset \cite{behley2019semantickitti}. We observed that there are no static motorcyclists or bicyclists in the training set. This means that using separate logits for moving and non-moving classes, as conventionally designed, will fail to identify any static bicyclists or motorcyclists, as there is no training data to learn from. To address this issue, we added a motion header to perform binary segmentation of moving and non-moving objects, and later fused it with the semantic header. By training the network to distinguish between moving and non-moving objects, such as pedestrians and vehicles, we aim to enable the network to recognize static and moving bicyclists and motorcyclists, even in the absence of any training data for those classes.
Specifically, we apply \textit{LogSoftMax} to normalize the motion logits and add them to each of the semantic logits that belong to movable classes. Consequently, we form moving and static logits for each of the movable logit. The implementation details are illustrated in Fig.~\ref{fig:motion_header}. The network is supervised by applying segmentation loss (\ie weighted combination of cross entropy, Lovasz softmax, and the proposed regularizer) on the motion logits, semantic logits and the final logits.

\begin{figure}[t]
  \centering
  \includegraphics[width=0.4\linewidth]{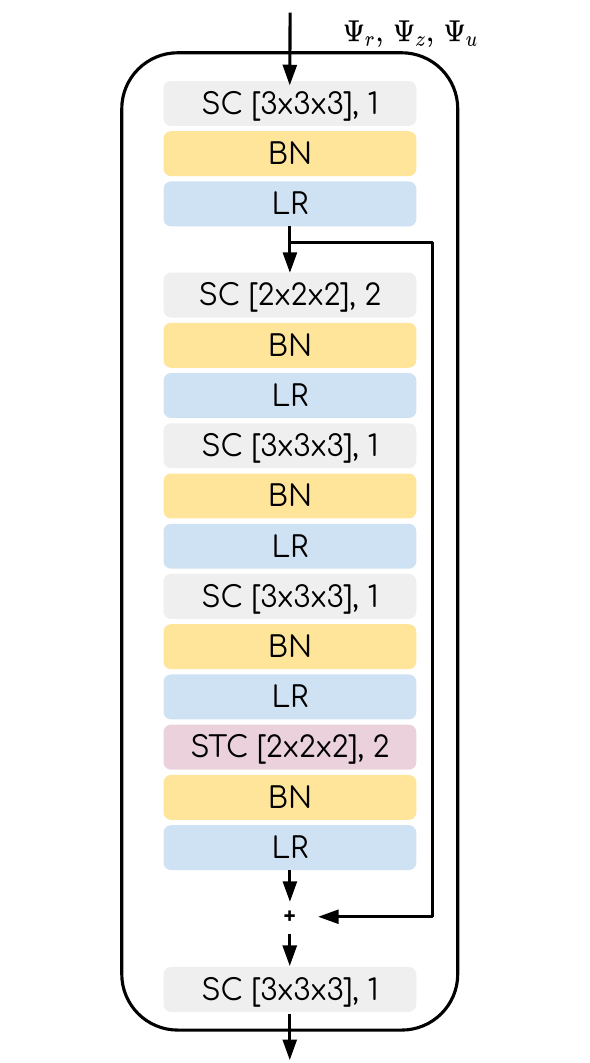}
  \caption{Illustration of the sparse convolutional blocks ($\Psi_r,\Psi_z,\Psi_u$) in MRM. SC: sparse 3D convolution [kenel size], stride. STC: 3D sparse transpose convolution [kernel size], stride. BN: BatchNorm. LR: LeakyReLU.}
    \label{fig:update_arch}
  \vspace{-10px}
\end{figure}

%% file: sections/supp_mrm.tex
\subsection{Details on Memory Refinement Module}
\label{sec:mrm}

Memory Refinement Module (MRM) is an improved version of ConvGRU \cite{ballas2016convgru} that updates the latent memory with the current observation embeddings as follows,
\begin{equation}
    \begin{aligned}
        r_t = \text{sigmoid}[\Psi_r( X'_{F,t}, H'_{F, t-1})], \\ 
        z_t = \text{sigmoid}[\Psi_z( X'_{F,t}, H'_{F, t-1})], \\
        \hat{H}_{F,t} = \text{tanh}[\Psi_u( X'_{F,t}, r_t \cdot H'_{F, t-1})], \\
        H_{F,t} = \hat{H}_{F,t} \cdot z_t + H'_{F, t-1} \cdot(1 - z_t), \\
    \end{aligned}
\end{equation}
where $X'_{F,t}$ is the current observation embeddings at time $t$, $H'_{F, t-1}$ is the latent memory embeddings at $t-1$, and $H_{F,t}$ is the updated latent memory. $\Psi_r, \Psi_z, \Psi_u$ are a single sparse 3D convolutional layer in the vanilla sparse ConvGRU \cite{ballas2016convgru}. However, we introduce a new design where they are implemented as sparse 3D convolutional blocks. These blocks integrate downsampling layers to expand the receptive field and upsampling layers to restore the embeddings to their original size. We provide a more detailed illustration of this design in Fig.~\ref{fig:update_arch}.

%% file: sections/supp_results.tex
\section{Additional Results}
\subsection{SemanticKITTI single-scan results}
\label{sec:sk_single}
Tab.~\ref{tab:skitti_single} compares \proposed \ with state-of-the-art approaches on the test set of SemanticKITTI single-scan benchmark. This is a more competitive benchmark focusing on single-scan semantic segmentation where previous research has focused on proposing various architectures \cite{cheng20212,xu2021rpvnet,zhu2021cylindrical} or knowledge distillation techniques \cite{hou2022pvkd}. Our results show that \proposed{} can still outperform these methods, which are highly optimized for this benchmark. Tab.~\ref{tab:skitti_single_val_sota} compares our apporach with the others on the validation set of the same benchmark. Please note that most prior works have only reported the mIoU metric on the validation set. Therefore, we only included comparison of mIoU in this table but presented detailed class-wise IoUs of our approach in Tab.~\ref{tab:sk_val}.
\begin{table*}[h]
    {
    \centering
    \resizebox{\textwidth}{!}{
    \begin{tabular}{l|c|ccccccccccccccccccc}
    \hline 
    \textbf{Method}  
    & \begin{sideways} \textbf{mIoU} \end{sideways} 
    & \begin{sideways} Car \end{sideways} 
    & \begin{sideways} Bicycle \end{sideways} 
    & \begin{sideways} Motorcycle \end{sideways} 
    & \begin{sideways} Truck \end{sideways} 
    & \begin{sideways} Other Vehicle \end{sideways} 
    & \begin{sideways} Person \end{sideways} 
    & \begin{sideways} Bicyclist \end{sideways} 
    & \begin{sideways} Motorcyclist \end{sideways} 
    & \begin{sideways} Road \end{sideways} 
    & \begin{sideways} Parking \end{sideways} 
    & \begin{sideways} Sidewalk \end{sideways} 
    & \begin{sideways} Other Ground \end{sideways} 
    & \begin{sideways} Building \end{sideways} 
    & \begin{sideways} Fence \end{sideways} 
    & \begin{sideways} Vegetation \end{sideways} 
    & \begin{sideways} Trunk \end{sideways} 
    & \begin{sideways} Terrain \end{sideways} 
    & \begin{sideways} Pole \end{sideways} 
    & \begin{sideways} Traffic Sign \end{sideways} \\ 
    \hline
    PointNet \cite{qi2017pointnet} & $14.6$ & $46.3$ & $1.3$ & $0.3$ & $0.1$ & $0.8$ & $0.2$ & $0.2$ & $0.0$ & $61.6$ & $15.8$ & $35.7$ & $1.4$ & $41.4$ & $12.9$ & $31.0$ & $4.6$ & $17.6$ & $2.4$ & $3.7$ \\
    RangeNet++ \cite{milioto2019rangenet++} & $52.2$ & $91.4$ & $25.7$ & $34.4$ & $25.7$ & $23.0$ & $38.3$ & $38.8$ & $4.8$ & $91.8$ & $65.0$ & $75.2$ & $27.8$ & $87.4$ & $58.6$ & $80.5$ & $55.1$ & $64.6$ & $47.9$ & $55.9$ \\
    RandLANet \cite{hu2020randla} & $53.9$ & $94.2$ & $26.0$ & $25.8$ & $40.1$ & $38.9$ & $49.2$ & $48.2$ & $7.2$ & $90.7$ & $60.3$ & $73.7$ & $20.4$ & $86.9$ & $56.3$ & $81.4$ & $61.3$ & $66.8$ & $49.2$ & $47.7$ \\
    PolarNet \cite{zhang2020polarnet} & $54.3$ & $93.8$ & $40.3$ & $30.1$ & $22.9$ & $28.5$ & $43.2$ & $40.2$ & $5.6$ & $90.8$ & $61.7$ & $74.4$ & $21.7$ & $90.0$ & $61.3$ & $84.0$ & $65.5$ & $67.8$ & $51.8$ & $57.5$ \\
    SqueezeSegv3 \cite{xu2020squeezesegv3} & $55.9$ & $92.5$ & $38.7$ & $36.5$ & $29.6$ & $33.0$ & $45.6$ & $46.2$ & $20.1$ & $91.7$ & $63.4$ & $74.8$ & $26.4$ & $89.0$ & $59.4$ & $82.0$ & $58.7$ & $65.4$ & $49.6$ & $58.9$ \\
    TemporalLidarSeg\cite{duerr2020lidar} & $58.2$ & $94.1$ & $50.0$ & $45.7$ & $28.1$ & $37.1$ & $56.8$ & $47.3$ & $9.2$ & $91.7$ & $60.1$ & $75.9$ & $27.0$ & $89.4$ & $63.3$ & $83.9$ & $64.6$ & $66.8$ & $53.6$ & $60.5$ \\
    KPConv \cite{thomas2019kpconv} & $58.8$ & $96.0$ & $30.2$ & $42.5$ & $33.4$ & $44.3$ & $61.5$ & $61.6$ & $11.8$ & $88.8$ & $61.3$ & $72.7$ & $31.6$ & $90.5$ & $64.2$ & $84.8$ & $69.2$ & $69.1$ & $56.4$ & $47.4$ \\
    SalsaNext \cite{cortinhal2020salsanext} & $59.5$ & $91.9$ & $48.3$ & $38.6$ & $38.9$ & $31.9$ & $60.2$ & $59.0$ & $19.4$ & $91.7$ & $63.7$ & $75.8$ & $29.1$ & $90.2$ & $64.2$ & $81.8$ & $63.6$ & $66.5$ & $54.3$ & $62.1$ \\
    Meta-RangeSeg \cite{wang2022meta} & 61.0 & 93.9 & 50.1 & 43.8 & 43.9 & 43.2 & 63.7 & 53.1 & 18.7 & 90.6 & 64.3 & 74.6 & 29.2 & 91.1 & 64.7 & 82.6 & 65.5 & 65.5 & 56.3 & 64.2 \\
    FusionNet \cite{zhang12356deep} & $61.3$ & $95.3$ & $47.5$ & $37.7$ & $41.8$ & $34.5$ & $59.5$ & $56.8$ & $11.9$ & $91.8$ & $68.8$ & $77.1$ & $30.8$ & $\mathbf{92.5}$ & $\mathbf{69.4}$ & $84.5$ & $69.8$ & $68.5$ & $60.4$ & $66.5$ \\
    TornadoNet \cite{gerdzhev2021tornadonet} & $63.1$ & $94.2$ & $55.7$ & $48.1$ & $40.0$ & $38.2$ & $63.6$ & $60.1$ & $34.9$ & $89.7$ & $66.3$ & $74.5$ & $28.7$ & $91.3$ & $65.6$ & $85.6$ & $67.0$ & $71.5$ & $58.0$ & $65.9$ \\
    SPVNAS \cite{tang2020searching} & $67.0$ & $97.2$ & $50.6$ & $50.4$ & $56.6$ & $58.0$ & $67.4$ & $67.1$ & $50.3$ & $90.2$ & $67.6$ & $75.4$ & $21.8$ & $91.6$ & $66.9$ & $86.1$ & $73.4$ & $71.0$ & $64.3$ & $67.3$ \\
    Cylinder3D \cite{zhu2021cylindrical} & $67.8$ & $97.1$ & $67.6$ & $64.0$ & $\mathbf{59.0}$ & $58.6$ & $73.9$ & $67.9$ & $36.0$ & $91.4$ & $65.1$ & $75.5$ & $32.3$ & $91.0$ & $66.5$ & $85.4$ & $71.8$ & $68.5$ & $62.6$ & $65.6$ \\
    (AF)2S3-Net \cite{cheng20212} & $69.7$ & $94.5$ & $65.4$ & $\mathbf{86.8}$ & $39.2$ & $41.1$ & $\mathbf{80.7}$ & $\mathbf{80.4}$ & $\mathbf{74.3}$ & $91.3$ & $68.8$ & $72.5$ & $\mathbf{53.5}$ & $87.9$ & $63.2$ & $70.2$ & $68.5$ & $53.7$ & $61.5$ & $\mathbf{71.0}$ \\
    RPVNet \cite{xu2021rpvnet} & $70.3$ & $\mathbf{97.6}$ & $\mathbf{68.4}$ & $68.7$ & $44.2$ & $\mathbf{61.1}$ & $\mathbf{75.9}$ & $74.4$ & $\mathbf{73.4}$ & $\mathbf{93.4}$ & $\mathbf{70.3}$ & $\mathbf{80.7}$ & $33.3$ & $\mathbf{93.5}$ & $\mathbf{72.1}$ & $\mathbf{86.5}$ & $\mathbf{75.1}$ & $\mathbf{71.7}$ & $64.8$ & $61.4$ \\
    PVKD \cite{hou2022pvkd} & $71.2$ & $97.0$ & $67.9$ & $\mathbf{69.3}$ & $53.5$ & $60.2$ & $75.1$ & $73.5$ & $50.5$ & $\mathbf{91.8}$ & $\mathbf{70.9}$ & $\mathbf{77.5}$ & $\mathbf{41.0}$ & $92.4$ & $69.4$ & $\mathbf{86.5}$ & $\mathbf{73.8}$ & $\mathbf{71.9}$ & $\mathbf{64.9}$ & $65.8$ \\
    \hline
    \proposed \ [ours] & $\mathbf{71.3}$ & $\mathbf{97.4}$ & $\mathbf{68.1}$ & 69.1 & $\mathbf{58.7}$ & $\mathbf{65.7}$ & 75.2 & $\mathbf{76.4}$ & 56.2 & 89.8 & 65.6 & 74.8 & 32.1 & 91.9 & 67.8 & 85.2 & 73.7 & 70.5 & $\mathbf{66.4}$ & $\mathbf{70.1}$ \\
    \hline
    \end{tabular}
    }
    \caption{Comparison to the state-of-the-art methods on the test set of SemanticKITTI \cite{behley2019semantickitti} single-scan benchmark. 
    We include LiDAR-only published approaches at the time of submission. Metrics are provided in [\%]. Top two entries of each classes are bolded. }
    \label{tab:skitti_single} }
    \end{table*} 
    \begin{table}[h!]
        \begin{center}
        \scalebox{0.7}
        {
        \begin{tabular}{l|ccc}
        \hline
        \textbf{Method} & \textbf{mIoU}\\
        \hline \hline
        RandLANet \cite{hu2020randla} & 57.1 \\
        PolarNet \cite{zhang2020polarnet} & 54.9 \\
        TornadoNet \cite{gerdzhev2021tornadonet} & 64.5 \\
        SPVNAS \cite{tang2020searching} & 64.7 \\
        Cylinder3D \cite{zhu2021cylindrical} & 65.9 \\
        PVKD \cite{hou2022pvkd} & 66.4 \\
        RPVNet \cite{xu2021rpvnet} & 69.6 \\
        \hline
        \proposed \ [ours] & 70.8 \\
        \proposed \ [ours] + TTA & $\mathbf{71.5}$ \\
        \hline
        \end{tabular}
        }
        \end{center}
        \caption{Comparison to the state-of-the-art methods on the validation set of SemanticKITTI \cite{behley2019semantickitti}. Metrics are provided in [\%].}
        \label{tab:skitti_single_val_sota}
        \end{table}

\subsection{SemanticKITTI multi-scan validation results}
\label{sec:sk_multi}
\begin{table*}[h]
    {
    \centering
    \resizebox{\textwidth}{!}{
    \begin{tabular}{l|c|ccccccccccccccccccccccccc}
    \hline 
    \textbf{Method}  
    & \begin{sideways} \textbf{mIoU} \end{sideways} 
    & \begin{sideways} Car \end{sideways} 
    & \begin{sideways} Bicycle \end{sideways} 
    & \begin{sideways} Motorcycle \end{sideways} 
    & \begin{sideways} Truck \end{sideways} 
    & \begin{sideways} Other Vehicle \end{sideways} 
    & \begin{sideways} Person \end{sideways} 
    & \begin{sideways} Bicyclist \end{sideways} 
    & \begin{sideways} Motorcyclist \end{sideways} 
    & \begin{sideways} Road \end{sideways} 
    & \begin{sideways} Parking \end{sideways} 
    & \begin{sideways} Sidewalk \end{sideways} 
    & \begin{sideways} Other Ground \end{sideways} 
    & \begin{sideways} Building \end{sideways} 
    & \begin{sideways} Fence \end{sideways} 
    & \begin{sideways} Vegetation \end{sideways} 
    & \begin{sideways} Trunk \end{sideways} 
    & \begin{sideways} Terrain \end{sideways} 
    & \begin{sideways} Pole \end{sideways} 
    & \begin{sideways} Traffic Sign \end{sideways} 
    & \begin{sideways} car (m) \end{sideways} 
    & \begin{sideways} bicyclist (m) \end{sideways} 
    & \begin{sideways} person (m) \end{sideways} 
    & \begin{sideways} motorcyclist (m) \end{sideways} 
    & \begin{sideways} other-vehicle (m) \end{sideways} 
    & \begin{sideways} truck (m) \end{sideways} \\ 
    \hline
    5FB & 53.5 & 95.4 & 56.0 & 75.6 & 74.5 & 53.5 & $\mathbf{25.5}$ & 0.0 & 0.0 & 92.7 & 46.6 & 78.8 & 0.5 & 90.1 & 58.6 & 89.2 & 72.8 & 75.3 & 66.2 & $\mathbf{53.7}$ & $\mathbf{78.6}$ & 87.2 & 64.6 & 0.0 & 3.4 & 0.0 \\
    \proposed & $\mathbf{58.5}$ & $\mathbf{95.9}$ & $\mathbf{64.8}$ & $\mathbf{86.2}$ & $\mathbf{96.3}$ & $\mathbf{66.4}$ & $23.7$ & $0.0$ & $0.0$ & $\mathbf{95.5}$ & $\mathbf{55.7}$ & $\mathbf{83.9}$ & $\mathbf{5.0}$ & $\mathbf{91.4}$ & $\mathbf{62.7}$ & $\mathbf{89.7}$ & $\mathbf{73.9}$ & $\mathbf{78.1}$ & $\mathbf{66.7}$ & $52.1$ & $72.7$ & $\mathbf{94.4}$ & $\mathbf{72.0}$ & $0.0$ & $\mathbf{35.7}$ & $0.0$ \\
    \hline
    \end{tabular}
    }
    \caption{Comparison to our 5-frame baseline (5FB) on the validation set of SemanticKITTI \cite{behley2019semantickitti}. Movable actors are further divided into moving and static. Metrics are provided in [\%].}
    \label{tab:sk_multi_val} }
    \end{table*} 
Tab.\ref{tab:sk_multi_val} compares \proposed \ with the 5-frame-baseline (5FB). The 5FB uses the same network architecture of \proposed \ but without the memory update module. Additionally, the input is 5 consecutive LiDAR scans projected to the most recent ego vehicle frame. In contrast, our approach processes only one scan at a time. The results indicate that \proposed \ significantly outperforms 5FB in almost all categories. The improvement is most prominent in the case of movable objects such as \textit{moving bicyclist} and \textit{other vehicle}. The 5FB can only reason about motion over a short time interval (i.e., 5 frames of data or approximately 0.5 seconds), as processing longer sequences all at once is computationally infeasible. However, \proposed \ employs a latent 3D memory to encode information from a more extended period, enabling the network to better comprehend the motion of moving actors.

\subsection{nuScenes validation results}
\label{sec:nusc_val}
\begin{table*}[h!]
    {
    \centering
    \resizebox{0.95\textwidth}{!}{
    \begin{tabular}{l|cc|cccccccccccccccc}
    \hline 
    \textbf{Method}  
    & \begin{sideways} \textbf{mIoU} \end{sideways} 
    & \begin{sideways} \textbf{FW-mIoU} \end{sideways} 
    & \begin{sideways} Barrier \end{sideways} 
    & \begin{sideways} Bicycle \end{sideways} 
    & \begin{sideways} Bus \end{sideways} 
    & \begin{sideways} Car \end{sideways} 
    & \begin{sideways} Construction \end{sideways} 
    & \begin{sideways} Motorcycle \end{sideways} 
    & \begin{sideways} Pedestrain \end{sideways} 
    & \begin{sideways} Traffic Cone \end{sideways} 
    & \begin{sideways} Trailer \end{sideways} 
    & \begin{sideways} Truck \end{sideways} 
    & \begin{sideways} Drivable \end{sideways} 
    & \begin{sideways} Other Flat \end{sideways} 
    & \begin{sideways} Sidewalk \end{sideways} 
    & \begin{sideways} Terrain \end{sideways} 
    & \begin{sideways} Manmade \end{sideways} 
    & \begin{sideways} Vegetation \end{sideways}  \\ 
    \hline
    RangeNet++ \cite{milioto2019rangenet++} & $65.5$ & $-$ & $66.0$ & $21.3$ & $77.2$ & $80.9$ & $30.2$ & $66.8$ & $69.6$ & $52.1$ & $54.2$ & $72.3$ & $94.1$ & $66.6$ & $63.5$ & $70.1$ & $83.1$ & $79.8$ \\
    PolarNet \cite{zhang2020polarnet} & $71.0$ & $-$ & $74.7$ & $28.2$ & $85.3$ & $90.9$ & $35.1$ & $77.5$ & $71.3$ & $58.8$ & $57.4$ & $76.1$ & $96.5$ & $71.1$ & $74.7$ & $74.0$ & $87.3$ & $85.7$ \\
    Salsanext \cite{cortinhal2020salsanext} & $72.2$ & $-$ & $74.8$ & $34.1$ & $85.9$ & $88.4$ & $42.2$ & $72.4$ & $72.2$ & $63.1$ & $61.3$ & $76.5$ & $96.0$ & $70.8$ & $71.2$ & $71.5$ & $86.7$ & $84.4$ \\
    Cylinder3D \cite{zhu2021cylindrical} & $76.1$ & $-$ &  $76.4$ & $40.3$ & $91.2$ & $\mathbf{93.8}$ & $51.3$ & $78.0$ & $78.9$ & $64.9$ & $62.1$ & $84.4$ & $96.8$ & $71.6$ & $\mathbf{76.4}$ & $75.4$ & $90.5$ & $87.4$ \\
    RPVNet \cite{xu2021rpvnet} & $77.6$ & $-$ & $78.2$ & $43.4$ & $92.7$ & $93.2$ & $49.0$ & $85.7$ & $80.5$ & $66.0$ & $66.9$ & $84.0$ & $96.9$ & $73.5$ & $75.9$ & $\mathbf{76.0}$ & $90.6$ & $88.9$ \\
    \hline
    SFB & 76.7 & 89.2 & 77.6 & 42.0 & 92.7 & 92.5 & 44.7 & 83.8 & 79.1 & 65.1 & 66.2 & 81.6 & 96.7 & 75.9 & 75.1 & 75.2 & 90.2 & 88.7 \\
    \proposed \ [ours] & $\mathbf{81.1}$ & $\mathbf{90.0}$ & $\mathbf{78.8}$ & $\mathbf{57.0}$ & $\mathbf{95.2}$ & 92.9 & $\mathbf{60.0}$ & $\mathbf{89.3}$ & $\mathbf{86.3}$ & $\mathbf{70.8}$ & $\mathbf{73.8}$ & $\mathbf{87.2}$ & $\mathbf{96.9}$ & $\mathbf{76.4}$ & 75.8 & 75.3 & $\mathbf{91.5}$ & $\mathbf{89.8}$ \\
    \hline
    \end{tabular}
    }
    \caption{Comparison to the state-of-the-art methods on the validation set of nuScenes \cite{caesar2020nuscenes} LiDAR semantic segmentation benchmark. We include LiDAR-only published approaches that report their validation mIoU. Metrics are provided in [\%].}
    \label{tab:nusc_val} }
    \end{table*} 
In Tab.\ref{tab:nusc_val}, we compare against state-of-the-art methods and our baseline on the validation set of nuScenes \cite{caesar2020nuscenes}. \proposed again outperforms all methods with the largest gains observed in smaller objects such as \textit{bicycle}, \textit{pedestrian}, and \textit{traffic cone}. Those are particularly challenging for semantic segmentation networks due to the sparsity of the point clouds in this dataset. Nonetheless, our method overcomes this limitation by leveraging a 3D latent memory to enhance semantic reasoning of the sparse points.

\begin{table*}[h!]
    {
    \centering
    \resizebox{\textwidth}{!}{
    \begin{tabular}{l|c|ccccccccccccccccccc}
    \hline 
    \textbf{Method}  
    & \begin{sideways} \textbf{mIoU} \end{sideways} 
    & \begin{sideways} Car \end{sideways} 
    & \begin{sideways} Bicycle \end{sideways} 
    & \begin{sideways} Motorcycle \end{sideways} 
    & \begin{sideways} Truck \end{sideways} 
    & \begin{sideways} Other Vehicle \end{sideways} 
    & \begin{sideways} Person \end{sideways} 
    & \begin{sideways} Bicyclist \end{sideways} 
    & \begin{sideways} Motorcyclist \end{sideways} 
    & \begin{sideways} Road \end{sideways} 
    & \begin{sideways} Parking \end{sideways} 
    & \begin{sideways} Sidewalk \end{sideways} 
    & \begin{sideways} Other Ground \end{sideways} 
    & \begin{sideways} Building \end{sideways} 
    & \begin{sideways} Fence \end{sideways} 
    & \begin{sideways} Vegetation \end{sideways} 
    & \begin{sideways} Trunk \end{sideways} 
    & \begin{sideways} Terrain \end{sideways} 
    & \begin{sideways} Pole \end{sideways} 
    & \begin{sideways} Traffic Sign \end{sideways} \\ 
    \hline
    SFB w/o cutMix & 66.2 & 96.0 & 54.2 & 76.7 & 78.5 & 53.5 & 71.2 & 92.0 & 0.7 & 94.5 & 49.3 & 82.2 & 3.8 & 91.0 & 63.8 & 88.7 & 71.2 & 76.0 & 64.6 & 49.4 \\
    SFB & 67.2 & 96.9 & 60.0 & 79.5 & 76.9 & 67.0 & 74.0 & 91.2 & 1.6 & 94.5 & 49.4 & 82.0 & 6.1 & 90.6 & 61.0 & 88.0 & 69.0 & 74.4 & 64.8 & 50.8 \\
    M1 & 69.5 & 97.5 & 62.4 & 88.0 & 79.1 & 74.4 & 83.7 & 93.2 & 2.5 & 95.3 & 55.7 & 83.5 & 3.3 & 90.4 & 61.8 & 88.1 & 69.5 & 74.7 & 64.9 & 52.2 \\
    M2 & 69.7 & 97.6 & 58.4 & 86.2 & 94.5 & 77.8 & 83.1 & 94.0 & 0.0 & 94.9 & 54.5 & 83.0 & 3.9 & 90.9 & 63.5 & 87.3 & 68.8 & 71.5 & 65.1 & 50.0 \\
    M3 & 69.7 & 97.3 & 59.4 & 84.9 & 82.5 & 73.7 & 83.1 & 93.7 & 8.9 & 94.9 & 50.1 & 82.4 & 4.6 & 90.8 & 61.6 & 89.2 & 70.8 & 77.2 & 66.0 & 52.6 \\
    M4 [ours] & 70.8 & 97.4 & 61.5 & 89.1 & 93.0 & 76.2 & 83.6 & 95.0 & 0.3 & 95.3 & 52.4 & 83.0 & 7.5 & 91.4 & 64.1 & 89.3 & 73.6 & 77.2 & 65.6 & 50.4 \\
    M4 [ours] + TTA & $\mathbf{71.5}$ & $97.9$ & $64.6$ & $89.3$ & $95.4$ & $81.9$ & $84.6$ & $95.2$ & $0.0$ & $95.6$ & $52.9$ & $83.9$ & $2.9$ & $91.3$ & $62.9$ & $89.7$ & $74.4$ & $77.8$ & $66.5$ & $51.8$ \\
    \hline
    \end{tabular}
    }
    \caption{Class-wise IoUs of our ablated methods on the validation set of SemanticKITTI \cite{behley2019semantickitti}. Metrics are provided in [\%].}
    \label{tab:sk_val} }
    \end{table*} 
    
        \begin{table*}[h!]
            {
            \centering
            \resizebox{\textwidth}{!}{
            \begin{tabular}{l|c|ccccccccccccccccccc}
            \hline 
            \textbf{Memory Vox Size}
            & \begin{sideways} \textbf{mIoU} \end{sideways} 
            & \begin{sideways} Car \end{sideways} 
            & \begin{sideways} Bicycle \end{sideways} 
            & \begin{sideways} Motorcycle \end{sideways} 
            & \begin{sideways} Truck \end{sideways} 
            & \begin{sideways} Other Vehicle \end{sideways} 
            & \begin{sideways} Person \end{sideways} 
            & \begin{sideways} Bicyclist \end{sideways} 
            & \begin{sideways} Motorcyclist \end{sideways} 
            & \begin{sideways} Road \end{sideways} 
            & \begin{sideways} Parking \end{sideways} 
            & \begin{sideways} Sidewalk \end{sideways} 
            & \begin{sideways} Other Ground \end{sideways} 
            & \begin{sideways} Building \end{sideways} 
            & \begin{sideways} Fence \end{sideways} 
            & \begin{sideways} Vegetation \end{sideways} 
            & \begin{sideways} Trunk \end{sideways} 
            & \begin{sideways} Terrain \end{sideways} 
            & \begin{sideways} Pole \end{sideways} 
            & \begin{sideways} Traffic Sign \end{sideways} \\ 
            \hline
            
            $v_m =$ 0.25 m & 70.2 & 97.1 & 65.4 & 89.4 & 90.3 & 69.2 & 78.1 & 94.9 & 2.4 & 95.2 & 53.9 & 83.1 & 4.9 & 91.3 & 63.2 & 89.3 & 69.8 & 77.1 & 65.3 & 53.7 \\
            $v_m =$ 0.5 m & $\mathbf{70.8}$ & 97.4 & 61.5 & 89.1 & 93.0 & 76.2 & 83.6 & 95.0 & 0.3 & 95.3 & 52.4 & 83.0 & 7.5 & 91.4 & 64.1 & 89.3 & 73.6 & 77.2 & 65.6 & 50.4 \\
            $v_m =$ 1.0 m & 70.1 & 97.2 & 59.2 & 84.8 & 91.5 & 74.2 & 79.4 & 93.7 & 0.3 & 95.1 & 55.4 & 82.5 & 14.5 & 90.7 & 61.3 & 88.7 & 71.5 & 76.1 & 64.2 & 50.9 \\
            $v_m =$ 2.0 m & 67.9 & 97.2 & 55.1 & 76.3 & 79.8 & 73.6 & 74.3 & 92.0 & 1.4 & 94.6 & 53.4 & 82.0 & 7.3 & 90.8 & 61.9 & 87.9 & 71.1 & 74.4 & 65.4 & 52.3 \\
            \hline
            \end{tabular}
            }
            \caption{Ablation results of different memory voxel size $v_m$ on the validation set of SemanticKITTI \cite{behley2019semantickitti}. Metrics are provided in [\%].}
            \label{tab:ab_mem_vox_size} }
            \end{table*} 

            \begin{table*}[h]
                {
                \centering
                \resizebox{\textwidth}{!}{
                \begin{tabular}{l|c|ccccccccccccccccccc}
                \hline 
                \textbf{APM Neighbours}
                & \begin{sideways} \textbf{mIoU} \end{sideways} 
                & \begin{sideways} Car \end{sideways} 
                & \begin{sideways} Bicycle \end{sideways} 
                & \begin{sideways} Motorcycle \end{sideways} 
                & \begin{sideways} Truck \end{sideways} 
                & \begin{sideways} Other Vehicle \end{sideways} 
                & \begin{sideways} Person \end{sideways} 
                & \begin{sideways} Bicyclist \end{sideways} 
                & \begin{sideways} Motorcyclist \end{sideways} 
                & \begin{sideways} Road \end{sideways} 
                & \begin{sideways} Parking \end{sideways} 
                & \begin{sideways} Sidewalk \end{sideways} 
                & \begin{sideways} Other Ground \end{sideways} 
                & \begin{sideways} Building \end{sideways} 
                & \begin{sideways} Fence \end{sideways} 
                & \begin{sideways} Vegetation \end{sideways} 
                & \begin{sideways} Trunk \end{sideways} 
                & \begin{sideways} Terrain \end{sideways} 
                & \begin{sideways} Pole \end{sideways} 
                & \begin{sideways} Traffic Sign \end{sideways} \\ 
                \hline
                k=3 & 69.7 & 97.2 & 61.3 & 84.8 & 90.8 & 73.5 & 74.5 & 92.8 & 3.5 & 95.2 & 52.2 & 82.9 & 6.1 & 91.6 & 64.2 & 88.4 & 73.4 & 75.0 & 65.2 & 52.1 \\
                k=5 & $\mathbf{70.8}$ & 97.4 & 61.5 & 89.1 & 93.0 & 76.2 & 83.6 & 95.0 & 0.3 & 95.3 & 52.4 & 83.0 & 7.5 & 91.4 & 64.1 & 89.3 & 73.6 & 77.2 & 65.6 & 50.4 \\
                k=8 & 70.0 & 97.3 & 59.7 & 84.8 & 93.7 & 75.1 & 79.9 & 94.0 & 0.4 & 94.9 & 54.2 & 82.6 & 6.4 & 91.4 & 63.0 & 88.7 & 72.0 & 75.8 & 64.9 & 51.7 \\
                \hline
                \end{tabular}
                }
                \caption{Ablation results of different padding neighbourhood sizes $k$ in APM on the validation set of SemanticKITTI \cite{behley2019semantickitti}. Metrics are provided in [\%].}
                \label{tab:ab_mem_padding_size} }
                \end{table*} 
\subsection{Ablations}
\label{sec:ablation}
Tab.\ref{tab:sk_val} presents the detailed class-wise IoUs of the model for ablation presented in the main text. Please note that we follow the semantic class mapping of the single-scan benchmark while conducting ablation analysis on SemanticKITTI. This is due to the significant class imbalance that arises when attempting to separate all movable actors into moving and static classes. For instance, the validation set will not include any moving trucks or moving motorcyclists, and there will be less than 1000 points of static bicyclists. This can lead to increased noise in the ablation results. Therefore, we maintain the 19 semantic classes during the ablation process to ensure more robust results.

\paragraph{Influence of memory voxel size}
Tab.~\ref{tab:ab_mem_vox_size} shows the results of our ablation experiments using different voxel sizes ($v_m$) to retain latent memory. We found that smaller object classes, such as \textit{bicycle} and \textit{motorcycle}, benefited from using smaller voxel sizes. However, using a large voxel size, such as 2m, resulted in much worse performance for these classes, possibly because it mixed different objects within the same voxel. Overall, using a memory voxel size of 0.5m produced the best results.

\paragraph{Influence of padding neighbourhood size in APM}
We present the results of our experiment on using different neighborhood sizes to aggregate embeddings for padding, as shown in Tab.~\ref{tab:ab_mem_padding_size}. Specifically, we tested neighborhood sizes of 3, 5, and 8 entries. We found that changing the padding neighborhood size had only a minimal effect on the background classes, which are typically static and do not move. This is because the closest entries usually have the most influence, so varying the neighborhood size had little impact. However, we observed more significant differences in the movable actors. For example, increasing the padding neighborhood size was most beneficial for the \textit{truck} class, where larger receptive fields are needed to aggregate potentially moving trucks. Overall, aggregating the closest 5 entries (k=5) produced the most favorable results.

\paragraph{Influence of instance cutMix}
The SemanticKITTI \cite{behley2019semantickitti} dataset contains many frames without movable actors, such as pedestrians and riders. The training set, on average, has only 0.63 pedestrians and 0.18 riders per frame. To address this significant class imbalance, we created an instance library that includes movable instances from the training sequences similar to \cite{zhou2021panoptic}. In each training iteration, we randomly select 5 instances from the library and add them to the scene. The sampling weight is determined by the inverse frequency of the class. This approach has been effective, as shown in Tab.~\ref{tab:sk_val}, resulting in an improvement from $66.2\%$ to $67.2\%$, with the most significant gains coming from the movable actors added during training.

\paragraph{Influence of test-time augmentation}
We follow existing works \cite{zhu2021cylindrical,hou2022pvkd} to apply test-time augmentation (TTA) for further improving the segmentation results. Specifically, we randomly sample an augmentation that includes rotation on the Z axis from $-\pi$ to $\pi$ and a global scaling factor ranging from $0.95$ to $1.05$. We apply this same augmentation to the entire sequence during inference and repeat the process 10 times, each with a different augmentation. We then average the prediction results from each of the 10 passes to obtain the final prediction.  Our experiments demonstrate that TTA improves the mIoU by $0.7\%$, with a slight improvement observed in every class IoU, as shown in the last row of Tab.~\ref{tab:sk_val}.

\subsection{Visualization of memory}
\label{sec:mem_vis}
\begin{figure*}[h!]
    \centering
    \includegraphics[width=\textwidth]{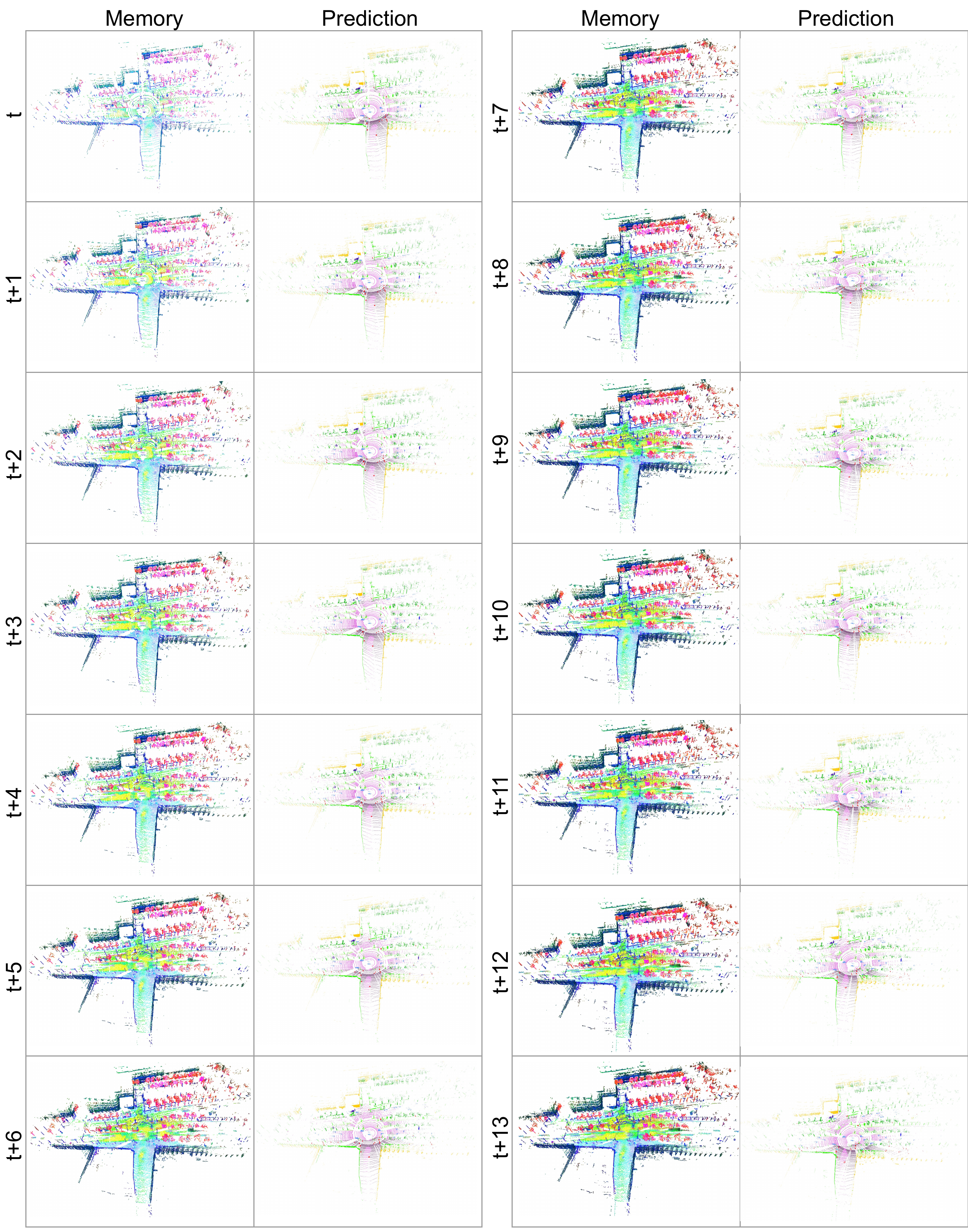}
    \caption{
    Illustration of the latent memory and prediction when unrolling the LiDAR sequence.
        }
    \label{fig:mem_qua}
\end{figure*}

We present a visualization of the 3D latent memory in Fig.~\ref{fig:mem_qua}, where PCA is used to reduce the embedding dimension to RGB. On the right side of the memory, we display the prediction generated by our network on the single scan. It is difficult to identify objects in a single scan due to the lack of semantic information and sparse observations, and occluded regions have no observations at all. In contrast, our latent memory is much denser, contains rich semantics that help separate different classes, and provides contextual details in occluded areas.

\subsection{Qualitative comparison}
\label{sec:qua}
\begin{figure*}[h!]
    \centering
    \includegraphics[width=\textwidth]{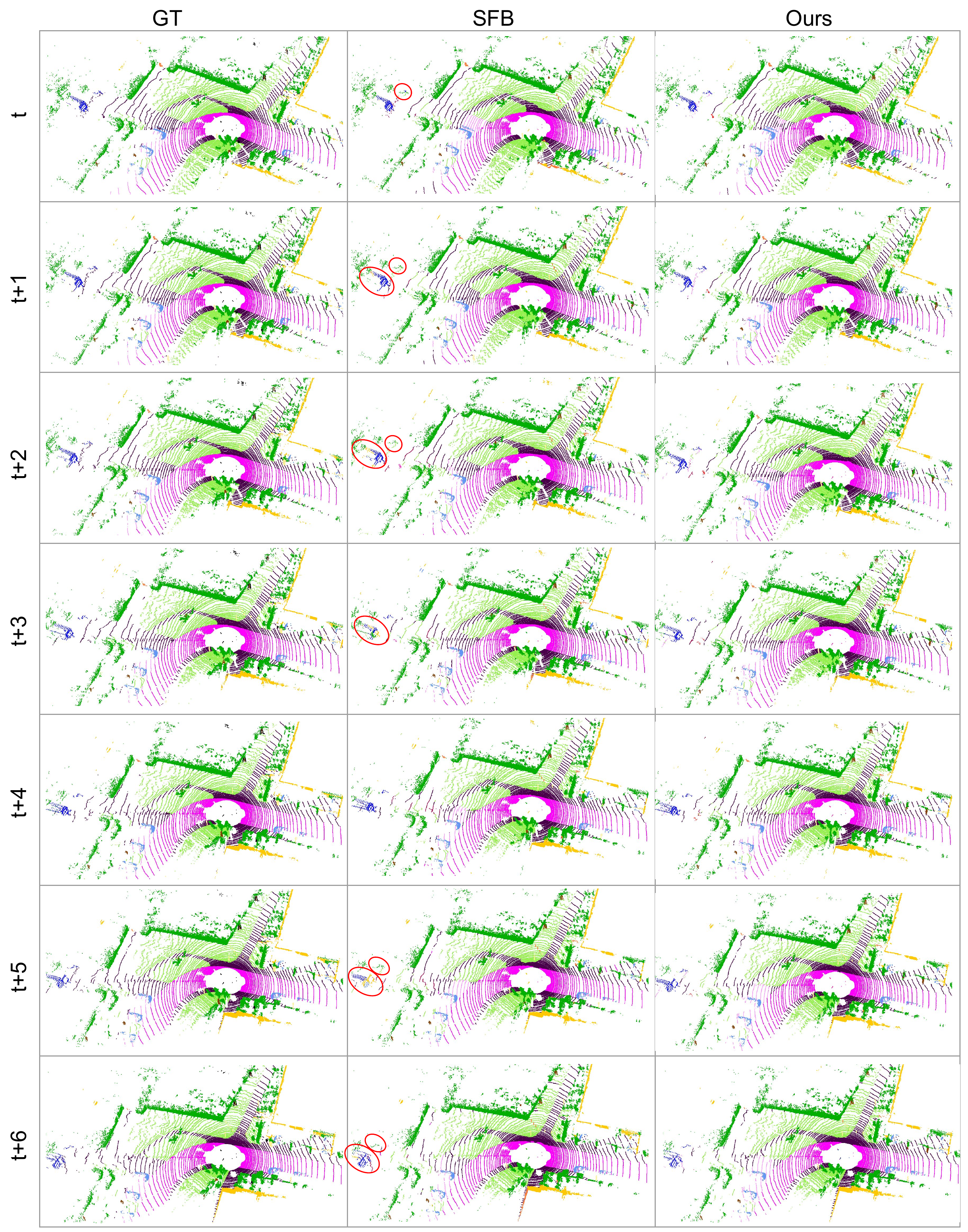}
    \caption{
    Qualitative comparison with single-frame baseline (SFB). Our approach is able to generate robust segmentation predictions throughout the interval where the SFB produces flickering results. See the vehicle highlighted in red circle.
        }
    \label{fig:sk_qua}
\end{figure*}

Fig.~\ref{fig:sk_qua} shows a qualitative comparison with our baseline. Please focus your attention to the two vehicles parked on the far left, highlighted with red circles. These scenarios are difficult for semantic segmentation because of the limited observations and partial occlusions. Despite these challenges, MemorySeg consistently segments the object accurately without any flickering. Conversely, the single-frame baseline fails to identify the parked vehicle in some frames, and the segmentation results fluctuate over time.

\begin{figure*}[h!]
    \centering
    \includegraphics[width=\textwidth]{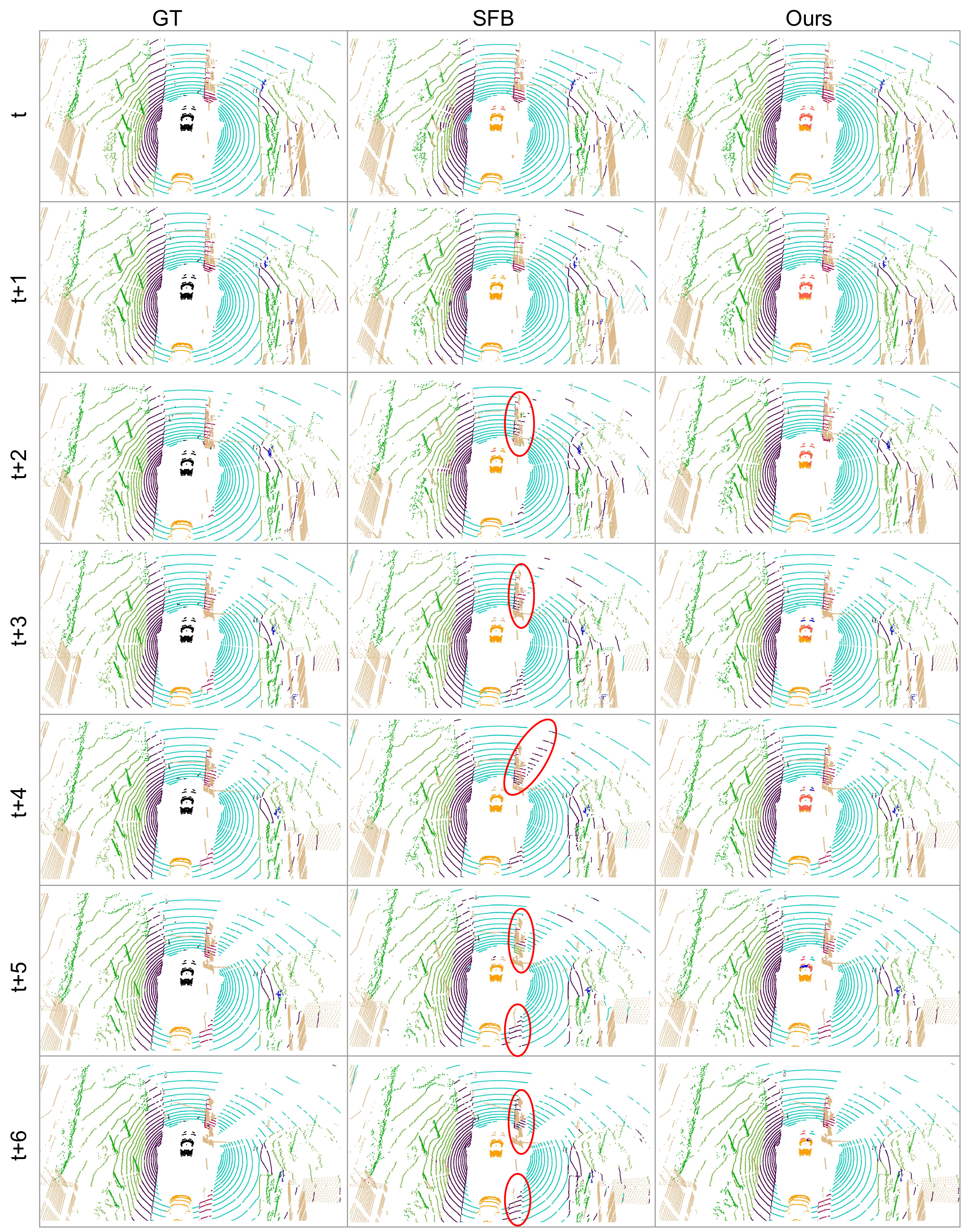}
    \caption{
    Qualitative comparison with single-frame baseline (SFB). SFB is prone to errors in regions with sparse observations or occlusions, resulting in confusion between sidewalks, roads, and other flat surfaces (highlighted in red). Our approach produces accurate and reliable results.
        }
    \label{fig:nusc_qua}
\end{figure*}

Furthermore, we present another qualitative example from the nuScenes \cite{caesar2020nuscenes} dataset in Fig.~\ref{fig:nusc_qua}, where we demonstrate substantial improvements in the background classes. Those classes often require an understanding of the surrounding environment to be segmented correctly. Our method improves contextual reasoning by accumulating past observations using a latent memory representation. Hence, while the single-frame baseline (SFB) is prone to errors, our approach yields accurate and reliable results.

